\RequirePackage{fix-cm}
\documentclass[twocolumn]{svjour3}          
\smartqed  
\usepackage{graphicx}
\usepackage[caption=false]{subfig}
\usepackage{multicol}
\usepackage{amsmath}
\usepackage{enumitem}
\usepackage{mathrsfs}
\usepackage{gensymb}
\usepackage{booktabs}
\usepackage{esvect}
\usepackage{color}
\usepackage[activate={true,nocompatibility},final,tracking=true,kerning=true,spacing=true,factor=1100,stretch=10,shrink=10]{microtype}
\usepackage{hyperref}
\journalname{International Journal of Computer Vision}
\makeatletter
\newcommand*{\rom}[1]{\expandafter\@slowromancap\romannumeral #1@}
\newcommand{\argmin}{\arg\!\min}
\begin{document}
%
\title{Temporally coherent general dynamic scene reconstruction}
\titlerunning{Temporally coherent general dynamic scene reconstruction} 

\author{Armin~Mustafa,
	Marco~Volino,
	Hansung~Kim,
	Jean-Yves~Guillemaut,
	and~Adrian~Hilton
}

\institute{All authors \at
	Centre for Vision, Speech and Signal Processing (CVSSP), University of Surrey, GU27XH, Guildford \\
	\email{a.mustafa, m.volino, h.kim, j.guillemaut and a.hilton @surrey.ac.uk}
}

\date{Received: date / Accepted: date}

\maketitle

\begin{abstract}
	Existing techniques for dynamic scene reconstruction from multiple wide-baseline cameras primarily focus on reconstruction in controlled environments, with fixed calibrated cameras and strong prior constraints.
	This paper introduces a general approach to obtain a 4D representation of complex dynamic scenes from multi-view wide-baseline static or moving cameras without prior knowledge of the scene structure, appearance, or illumination.
	Contributions of the work are: An automatic method for initial coarse reconstruction to initialize joint estimation; Sparse-to-dense temporal correspondence integrated with joint multi-view segmentation and reconstruction to introduce temporal coherence; and a general robust approach for joint segmentation refinement and dense reconstruction of dynamic scenes by introducing shape constraint.
	Comparison with state-of-the-art approaches on a variety of complex indoor and outdoor scenes, demonstrates improved accuracy in both multi-view segmentation and dense reconstruction. 
	This paper demonstrates unsupervised reconstruction of complete temporally coherent 4D scene models with improved non-rigid object segmentation and shape reconstruction and its application to various applications such as free-view rendering and virtual reality.
	\keywords{Dynamic 4D reconstruction, Segmentation}
	\end{abstract}
\begin{figure}[t]
	\begin{center}
		\includegraphics[width = 0.99\linewidth]{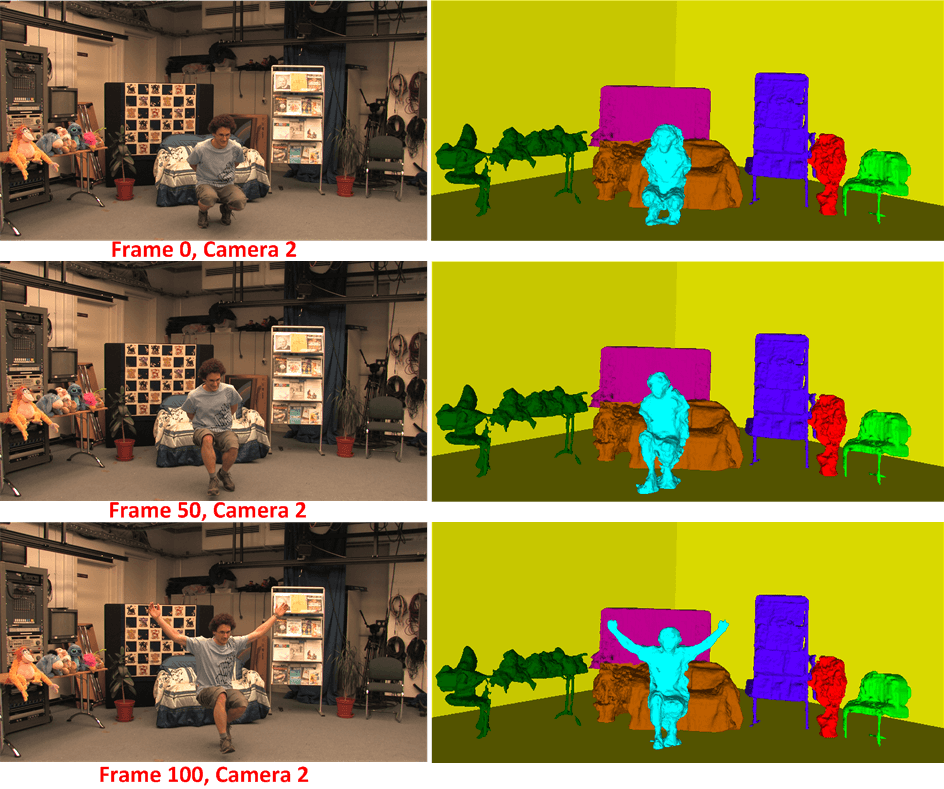}
		\caption{Temporally consistent color-coded scene reconstruction and segmentation for Odzemok dataset.}
		\label{fig:motivationCVPR}
	\end{center}
\end{figure}
\section{Introduction}\label{sec:introduction}
Reconstruction of general dynamic scenes is of great importance in entertainment applications such as visual effects in film and broadcast production and for content production in virtual reality. The ultimate goal of modelling dynamic scenes from multiple cameras is automatic understanding of real-world scenes from distributed camera networks, for applications in robotics and other autonomous systems.
Existing multi-view dynamic scene reconstruction methods either work in controlled environment with known background or chroma-key studio \cite{Guillemaut2010,Goldluecke04,Starck2007,taneja2011modeling} or require a large number of cameras \cite{Furukawa2010}. Extensions to more general outdoor scenes \cite{UnstructuredVBR10,Kim2012,taneja2011modeling} use prior reconstruction of the static geometry from images of the empty environment.
However these methods either require accurate segmentation of dynamic foreground objects, or prior knowledge of the scene structure and background, or are limited to static cameras and controlled environments. 
Scenes are reconstructed semi-automatically, requiring manual intervention for segmentation/rotoscoping, and result in temporally incoherent per-frame mesh geometries. Temporally coherent geometry with known surface correspondence across the sequence is essential for real-world applications and compact representation.

This paper addresses the limitations of existing approaches by introducing a methodology for unsupervised 4D temporally coherent dynamic scene reconstruction from multiple wide-baseline static or moving camera views without prior knowledge of the scene structure or background appearance.
The scene is automatically decomposed into a set of spatio-temporally coherent objects as shown in Figure \ref{fig:motivationCVPR} where the resulting 4D scene reconstruction has temporally coherent labels and surface correspondence for each object. This temporally coherent dynamic scene reconstruction is demonstrated to work in applications for immersive content production such as free-viewpoint video (FVV) and virtual reality (VR).
The contributions are summarized as follows:
\begin{itemize}[topsep=0pt,partopsep=0pt,itemsep=0pt,parsep=0pt] 
	\item Unsupervised temporally coherent dense reconstruction and segmentation of general complex dynamic scenes from multiple wide-baseline views.
	\item Automatic initialization of dynamic object segmentation and reconstruction from sparse features.
	\item A framework for space-time sparse-to-dense segmentation, reconstruction and temporal correspondence.
	\item Robust spatio-temporal refinement of dense reconstruction and segmentation integrating error tolerant photo-consistency and edge information using geodesic star convexity.
	\item Robust and computationally efficient reconstruction of dynamic scenes by exploiting temporal coherence.
	\item Real-world applications of 4D reconstruction to free-viewpoint video rendering and virtual reality. 
\end{itemize}
This paper presents a unified framework from two previously published papers, combining multiple view joint reconstruction and segmentation \cite{MustafaICCV15} with temporal coherence \cite{Mustafa16} to improve per-frame reconstruction performance. In particular the approach estimates a 4D surface model with full correspondence over time. A comprehensive experimental evaluation with comparison to the state-of-the-art in segmentation, reconstruction and 4D modelling is also presented extending previous work. Application of the resulting 4D models to FVV rendering and content production for immersive VR experiences is also presented.
%
\section{Related Work}
Temporally coherent reconstruction is a challenging task for general dynamic scenes due to a number of factors such as motion blur, articulated, non-rigid and large motion of multiple people, resolution differences between camera views, occlusions, wide-baselines, errors in calibration and cluttered dynamic backgrounds. 
Segmentation of dynamic objects from such scenes is difficult because of foreground and background complexity and the likelihood of overlapping background and foreground color distributions. 
Reconstruction is also challenging due to limited visual cues and relatively large errors affecting both calibration and extraction of a globally consistent solution.
This section reviews previous work on dynamic scene reconstruction and segmentation.
\subsection{Dynamic Scene Reconstruction}
Dense dynamic shape reconstruction is a fundamental problem and heavily studied area in the field of computer vision. Recovering accurate 3D models of a dynamically evolving, non-rigid scene observed by multiple synchronised cameras is a challenging task.
Research on multiple view dense dynamic reconstruction has primarily focused on indoor scenes with controlled illumination and static backgrounds, extending methods for multiple view reconstruction of static scenes \cite{Seitz06} to sequences \cite{Tung09}. Deep learning based approaches have been introduced to estimate shape of dynamic objects from minimal camera views in constrained environment \cite{Huang_2018_ECCV,Wu_2018_ECCV} and for rigid objects \cite{Stutz2018}.
In the last decade, focus has shifted to more challenging outdoor scenes captured with both static and moving cameras.
Reconstruction of non-rigid dynamic objects in uncontrolled natural environments is challenging due to the scene complexity, illumination changes, shadows, occlusion and dynamic backgrounds with clutter such as trees or people.
Methods have been proposed for multi-view reconstruction \cite{Vo_2016_CVPR,Cheng09,Larsen07} requiring a large number of closely spaced cameras for surface estimation of dynamic shape. 
Practical applications require relatively sparse moving cameras to acquire coverage over large areas such as outdoor.
A number of approaches for mutli-view reconstruction of outdoor scenes require initial silhouette segmentation \cite{Wu2013,Kim2012,Li10,Guillemaut2010} to allow visual-hull reconstruction. 
Most of these approaches to general dynamic scene reconstruction fail in the case of complex (cluttered) scenes captured with moving cameras. 

A recent work proposed reconstruction of dynamic fluids \cite{Qian_2017_CVPR} for static cameras. Another work used RGB-D cameras to obtain reconstruction of non-rigid surfaces \cite{Slavcheva_2017_CVPR}.
Pioneering research in general dynamic scene reconstruction from multiple handheld wide-baseline cameras \cite{UnstructuredVBR10,taneja2011modeling} exploited prior reconstruction of the background scene to allow dynamic foreground segmentation and reconstruction. Recent work \cite{Ngo2019} estimates shape of dynamic objects from handheld cameras exploiting GANs.
However these approaches either work for static/indoor scenes or exploit strong prior assumptions such as silhouette information, known background or scene structure. Also all these approaches give per frame reconstruction leading to temporally incoherent geometries.
Our aim is to perform temporally coherent dense reconstruction of unknown dynamic non-rigid scenes automatically without strong priors or limitations on scene structure.
\subsection{Multi-view Video Segmentation}
In the field of image segmentation, approaches have been proposed to provide temporally consistent monocular video segmentation \cite{GrundmannKwatra2010,Papazoglou13,Narayana13,Zhang13}. Hierarchical segmentation based on graphs was proposed in \cite{GrundmannKwatra2010}, directed acyclic graph was used to propose an object segmentation \cite{Zhang13} and optical flow is used to consistently segment objects \cite{Narayana13,Papazoglou13}.
Recently a number of approaches have been proposed for multi-view foreground object segmentation by exploiting appearance similarity spatially across views \cite{Djelouah13,Kowdle12,Lee11,Zeng04} and space-time similarity \cite{Djelouah15}. 
However, multi-view approaches assume a static background and different colour distributions for the foreground and background which limits applicability for general scenes and non-rigid objects.

To address this issue we introduce a novel method for spatio-temporal multi-view segmentation of dynamic scenes using shape constraints. Single image segmentation techniques using shape constraints provide good results for complex scenes \cite{Gulshan10} (convex and concave shapes), but require manual interaction. The proposed approach performs automatic multi-view video segmentation by initializing the foreground object model using spatio-temporal information from wide-baseline feature correspondence followed by a multi-layer optimization framework. Geodesic star convexity previously used in single view segmentation \cite{Gulshan10} is applied to constraint the segmentation in each view. 
Our multi-view formulation naturally enforces coherent segmentation between views and resolves ambiguities such as the similarity of background/foreground in isolated views.
%
\subsection{Joint Segmentation and Reconstruction}
Joint segmentation and reconstruction methods incorporate estimation of segmentation or matting with reconstruction to provide a combined solution. Joint refinement avoids the propagation of errors between the two stages thereby making the solution more robust. Also, cues from segmentation and reconstruction can be combined efficiently to achieve more accurate results.
The first multi-view joint estimation system was proposed by Szeliski et al.\cite{Szeliski99} which used iterative gradient descent to perform an energy minimization.
A number of approaches were introduced for joint formulation in static scenes and one recent work used training data to classify the segments \cite{Zach13joint3d}. The focus shifted to joint segmentation and reconstruction for rigid objects in indoor and outdoor environments. These approaches used a variety of techniques such as patch-based refinement \cite{Shin2013,Ozden07} and fixating cameras on the object of interest \cite{Campbell201014} for reconstructing rigid objects in the scene. However, these are either limited to static scenes \cite{Zach13joint3d,Hane13} or process each frame independently thereby failing to enforce temporal consistency \cite{Campbell201014,Guillemaut2010}.

Joint reconstruction and segmentation on monocular video was proposed in \cite{Abhijit14,Abarghouei_2019_CVPR,Chen_2019_CVPR} achieving semantic segmentation of scene limited to rigid objects in street scenes. Practical application of joint estimation requires these approaches to work on non-rigid objects such as humans with clothing.
A multi-layer joint segmentation and reconstruction approach was proposed for multiple view video of sports and indoor scenes \cite{Guillemaut2010}. The algorithm used known background images of the scene without the dynamic foreground objects to obtain an initial segmentation. 
Visual-hull based reconstruction was performed with known prior foreground/background using a background image plate with fixed and calibrated cameras. This visual hull was used as a prior and was optimized by a combination of photo-consistency, silhouette, color and sparse feature information in an energy minimization framework to improve the segmentation and reconstruction quality. Although structurally similar to our approach, it requires the scene to be captured by fixed calibrated cameras and a priori known fixed background plate as a prior to estimate the initial visual hull by background subtraction. The proposed approach overcomes these limitations allowing moving cameras and unknown scene backgrounds.
These methods are able to produce high quality results, but rely on good initializations and strong prior assumptions with known and controlled (static) scene backgrounds.

To overcome the limitations of existing methods, the proposed approach automatically initialises the foreground object segmentation from wide-baseline correspondence without prior knowledge of the scene. This is followed by a joint spatio-temporal reconstruction and segmentation of general scenes.
%
%
\begin{figure*}[t]
	\begin{center}
		\includegraphics[width = 0.99\linewidth]{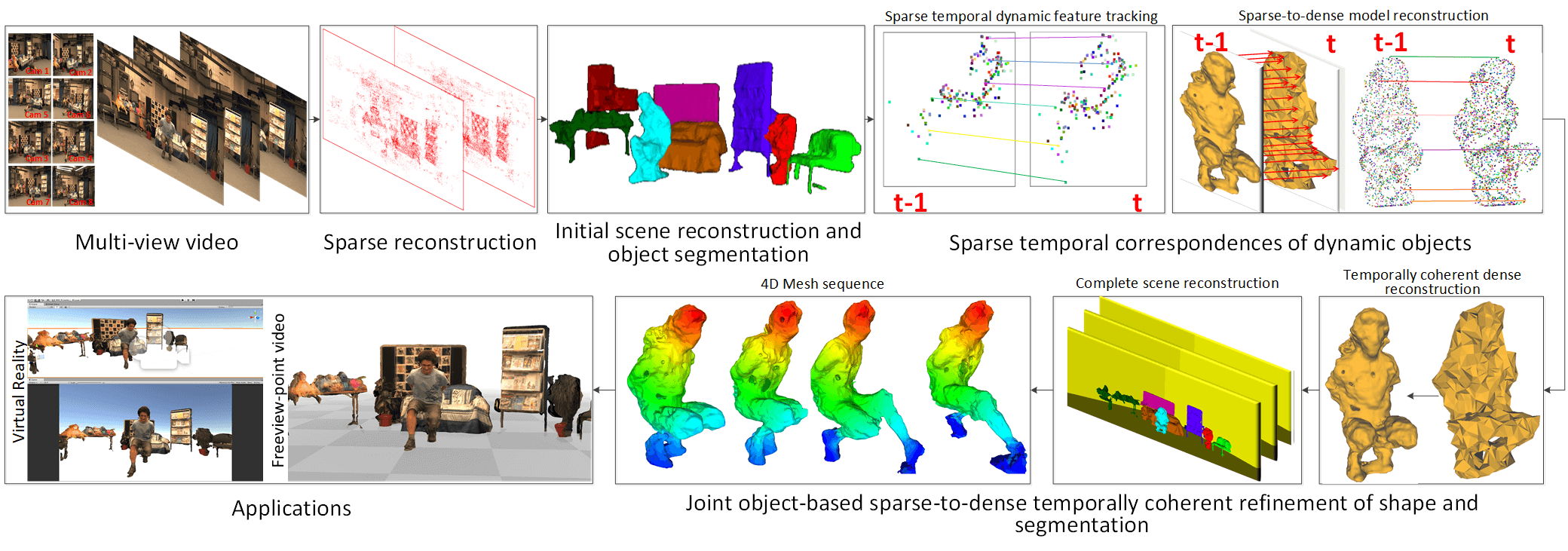}
		\vspace{-0.2cm}
		\caption{Overview of temporally consistent scene reconstruction framework.}
		\vspace{-0.75cm}
		\label{fig:algorithm}
	\end{center}
\end{figure*}
%
\vspace{-0.4cm}
\subsection{Temporally coherent 4D Reconstruction}
\vspace{-0.1cm}
Temporally coherent 4D reconstruction refers to aligning the 3D surfaces of non-rigid objects in time for a dynamic sequence. This is achieved by estimating point-to-point correspondences for the 3D surfaces to obtain temporally coherent 4D reconstruction. 4D models allows to create efficient representation for practical applications in film, broadcast and immersive content production such as virtual, augmented and mixed reality.
The existing approaches for multi-view reconstruction of dynamic scenes process each time frame independently. Independent per-frame reconstruction can result in errors due to the inherent visual ambiguity caused by occlusion and similar object appearance for general scenes.
3D scene flow \cite{Menze2015CVPR} estimates frame-to-frame correspondence or exploits 2D optical flow \cite{Wedel2011,Basha2013}. These methods require an accurate estimate for most of the pixels which fails in the case of large motion. However, 3D scene flow methods align two frames independently and do not produce temporally coherent 4D models across the complete sequence to obtain a single surface model.
Research investigating spatio-temporal reconstruction across multiple frames was proposed by \cite{Goldluecke04,Larsen07,Guillemaut3dv,Mustafa16ECCV} exploiting the temporal information from the previous frames using optical flow. An approach for recovering space-time consistent depth maps from multiple video sequences captured by stationary, synchronized and calibrated cameras for depth based free viewpoint video rendering was proposed by \cite{Cheng09}. However these methods require accurate initialisation, fixed and calibrated cameras and are limited to simple scenes.
Other approaches to temporally coherent reconstruction either require a large number of closely spaced cameras \cite{Bailer15} or bi-layer segmentation \cite{Zhang11robustbilayer,Jiang12} as a constraint for reconstruction. 
Recent approaches for spatio-temporal reconstruction of multi-view data work on indoor data \cite{Oswald14}.

The proposed framework addresses limitations of existing approaches and gives 4D temporally coherent reconstruction for general dynamic indoor or outdoor scenes with large non-rigid motions, repetitive texture, uncontrolled illumination, and large capture volume. The proposed approach gives 4D models of complete scenes with both static and dynamic objects for real-world applications (FVV and VR) with no prior knowledge of scene structure.
%
%
\subsection{Summary and Motivation}
Image-based temporally coherent 4D dynamic scene reconstruction without constraints is a key problem in computer vision. Existing dense reconstruction algorithms need some strong initial prior and constraints for the solution to converge such as background, structure, and segmentation, which limits their application for automatic reconstruction of general scenes. 
Current approaches are also commonly limited to independent per-frame reconstruction and do not exploit 
temporal information or produce a 4D model with known correspondence.
The approach proposed in this paper aims to overcome the limitations of existing approaches by initializing the joint reconstruction and segmentation algorithm automatically, introducing temporal coherence in the reconstruction and geodesic star convexity in segmentation to reduce ambiguity and ensure consistent non-rigid structure initialization at successive frames.
%
%
%
\section{Methodology}
\label{sec:method}
%
%
%
%
An overview of the proposed framework for temporally coherent multi-view reconstruction is presented in Figure \ref{fig:algorithm} and consists of the following stages:\\
\textbf{Multi-view video:}
The scenes are captured using multiple video cameras (static/moving) separated by wide-baseline ($>15^{\circ}$). Cameras can be synchronised either directly or in post-processing using the audio information. The cameras can be synchronized using time-code generator or later using the audio information. Camera intrinsics are known. Camera extrinsics (location,orientation) and scene structure are assumed to be unknown.\\
\textbf{Sparse reconstruction:}
Segmentation based feature detection (SFD) \cite{Mustafa15,Mustafa2019MSFDMS} is used to detect sparse features distributed throughout the scene for wide-baseline matching. SFD features are matched between views using a SIFT descriptor. The camera extrinsics and sparse 3D feature locations are then estimated for each time instant for the entire sequence \cite{Hartley03}.\\
\textbf{Initial coarse reconstruction - Section \ref{sec:sparseClustering}:}
The sparse point cloud is clustered in 3D \cite{RusuDoctoralDissertation} with each cluster representing a unique foreground object. Automatic initialisation is performed without prior knowledge of the scene structure or appearance to obtain an initial approximation for each object.\\
\textbf{Sparse-to-dense temporal reconstruction - Section \ref{sec:sparsecorrespondence}}
Temporal coherence is introduced to initialize the coarse reconstruction and obtain frame-to-frame dense correspondences for dynamic objects. Sparse temporal correspondence helps in identifying dynamic objects and allows propagation of the dense reconstruction between time instants to obtain an initialization.\\
\textbf{Joint refinement of shape and segmentation - Section \ref{sec:4Doptimize}:}
The initial estimate is refined for each object per-view through joint optimisation of shape and segmentation using a robust cost function combining matching, color, contrast and smoothness information with a geodesic star convexity constraint.
A single 3D model for each dynamic object is obtained by fusion of the view-dependent depth maps using Poisson surface reconstruction \cite{Kazhdan2006}. Surface orientation is estimated based on neighbouring pixels.
%
%
%
\subsection{Initial Coarse Reconstruction}
\label{sec:sparseClustering}
\begin{figure}[t]
	\begin{center}
		\includegraphics[width = 0.99\linewidth]{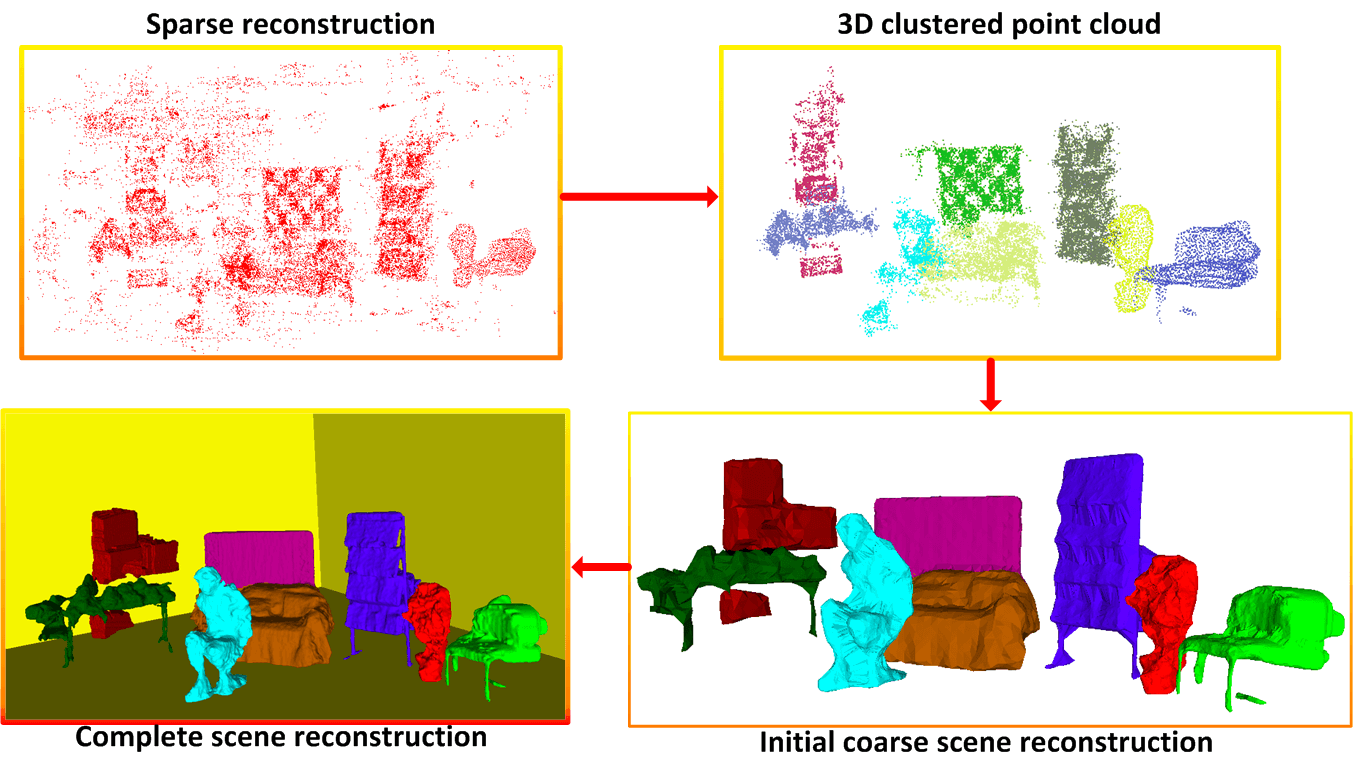}
		\caption{Overview of stages for estimation of an initial dense scene reconstruction. For more details refer to Section \ref{sec:sparseClustering}.}
		\label{fig:clustering}
	\end{center}
\end{figure}

For general dynamic scene reconstruction, we need to reconstruct and segment the objects in the scene. This requires an initial coarse approximation for initialisation of a subsequent refinement step to optimise the segmentation and reconstruction. Sparse point cloud clustering is used to segment objects, an overview is shown in Figure \ref{fig:clustering}.
The dense reconstruction of the foreground objects and background are combined to obtain a full scene reconstruction at the first time instant. A coarse geometric proxy of the background is created. For consecutive time instants dynamic objects and newly appeared objects are identified and only these objects are reconstructed and segmented. The reconstruction of static objects is retained which reduces computational complexity.
The optic flow and cluster information for each dynamic object ensures that we retain consistent labels for the entire sequence.
%
\subsubsection{Background Reconstruction} \label{sec:background}
Accurate reconstruction of the background is often challenging due to uniform appearance of large regions. A coarse geometric proxy of the background is created by computing the minimum oriented bounding box for the sparse 3D point-cloud using principal component analysis \cite{Dimitrov06}. 
Different methods are used for background estimation for indoor and outdoor scenes. 
For outdoor scenes a plane is inserted at infinity perpendicular to the ground plane as there are no consistent constraints.
For indoor scenes the Manhattan world assumption \cite{CoughlanY00} is applied to estimate room structure. The process used for estimation of the background is described below:
\begin{itemize}[topsep=0pt,partopsep=0pt,itemsep=0pt,parsep=0pt] 
	\item The centroid \textbf{A} $= (a_0, a_1, a_2)$ and normalized covariance of the point-cloud are estimated to compute the eigenvectors $\vv{e} = (e_{0}, e_{1}, e_{2})$ for the covariance matrix of the point-cloud. We define the reference system as \textbf{R}$ = (e_{0}, e_{1}, e_{0} \times e_{1})$ such that: $e_{0} \times e_{1} = \pm e_{2}$. Rotation matrix \textbf{R} and translation \textbf{A} are used to map sparse points in the first frame and place a box in correct location.
	\item The minimum and maximum values of coordinates in the x, y and z directions for the transformed cloud are computed to determine the minimum oriented box width, height, and depth.
	\item Given a box centred at the origin with size defined above the rotation \textbf{R} and translation \textbf{R}$\times$\textbf{C}$+$\textbf{A} is applied, where \textbf{C} is the middle of the minimum and maximum points.
\end{itemize}
This background reconstruction is a rough geometric proxy estimate of the background of the scene but gives reasonable results for complete scene reconstruction. 
%
\subsubsection{Sparse Point-cloud Clustering}
The sparse representation of the scene is processed to remove outliers using the point neighbourhood statistics to filter outlier data \cite{RusuDoctoralDissertation}. 
We segment the objects in the sparse scene reconstruction, this allows only moving objects to be reconstructed at each frame for efficiency and this also allows object shape similarity to be propagated across frames to increase robustness of reconstruction. Object segmentation increases efficiency and improve robustness of 4D models.

We use data clustering approach based on the 3D grid subdivision of the space using an octree data structure in Euclidean space to segment objects at each frame \cite{RusuDoctoralDissertation}. In a more general sense, nearest neighbor information is used to cluster, which is essentially similar to a flood fill algorithm. 
We choose this data clustering because of its computational efficiency and robustness. The approach allows segmentation of objects in the scene and is demonstrated to work well for cluttered and general outdoor scenes as shown in Section \ref{sec:results}.

Objects with insufficient detected features are reconstructed as part of the scene background. Appearing, disappearing and reappearing objects are handled by sparse dynamic feature tracking, explained in Section \ref {sec:sparsecorrespondence}. Clustering results are shown in Figure \ref{fig:clustering}. This is followed by a sparse-to-dense coarse object based approach to segment and reconstruct general dynamic scenes.
\begin{figure}[t]
	\begin{center}
		\includegraphics[width = 0.99\linewidth]{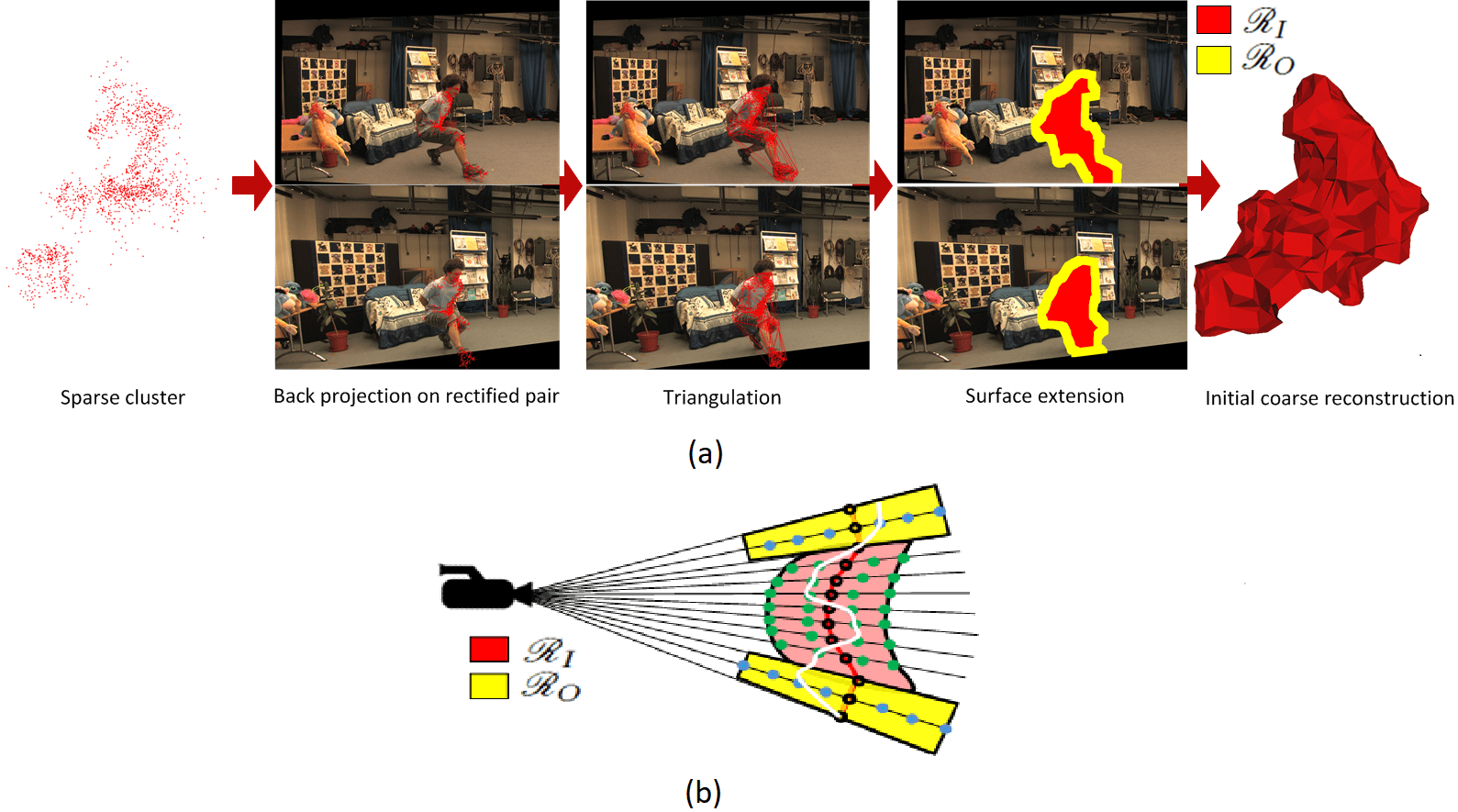}
		\caption{Initial coarse reconstruction: (a) Sparse-to-dense initial coarse reconstruction of the dynamic object in Odzemok dataset; and (b) White line represents the actual surface, Depth labels are represented as circles; blue circles depict depth labels in $\mathscr{D}_{O}$, green circles depict depth labels in $\mathscr{D}_{I}$ and black circles depict the initial surface estimate.}
		\label{fig:ICR}
	\end{center}
\end{figure}
\subsubsection{Coarse Object Reconstruction}
The process to obtain the coarse reconstruction for the first frame of the sequence is shown in Figure \ref{fig:ICR}. The sparse representation of each element is back-projected on the rectified image pair for each view. Delaunay triangulation \cite{Fortune97} is performed on the set of back projected points for each cluster on one image and is propagated to the second image using the sparse matched features. Triangles with edge length greater than the median length of edges of all triangles are removed. For each remaining triangle pair direct linear transform is used to estimate the affine homography. Displacement at each pixel within the triangle pair is estimated by interpolation to get an initial dense disparity map for each cluster in the 2D image pair labelled as $\mathscr{R}_{I}$ depicted in red in Figure \ref{fig:ICR}. The initial coarse reconstruction for the observed objects in the scene is used to define the depth hypotheses at each pixel for the optimization.

The region $\mathscr{R}_{I}$ does not ensure complete coverage of the object, so we extrapolate this region to obtain a region $\mathscr{R}_{O}$ (shown in yellow) in 2D by $5\%$ of the average distance between the boundary points($\mathscr{R}_{I}$) and the centroid of the object. 
To allow for errors in the initial approximate depth from sparse features we add volume in front and behind of the projected surface by an error tolerance, along the optical ray of the camera. This ensures that the object boundaries lie within the extrapolated initial coarse estimate. The tolerance for extrapolation may vary if a pixel belongs to $\mathscr{R}_{I}$ or $\mathscr{R}_{O}$ as the propagated pixels of the extrapolated regions ($\mathscr{R}_{O}$) may have a high level of error compared to error at the points from sparse representation ($\mathscr{R}_{I}$) requiring a comparatively higher tolerance. The calculation of threshold depends on the capture volume of the datasets and is set to $1\%$ of the capture volume for $\mathscr{R}_{O}$ and half the value for $\mathscr{R}_{I}$. This volume in 3D corresponds to our initial coarse reconstruction of each object and enables us to remove the dependency of previous approaches on static background plate and visual hull estimates.
This process of cluster identification and initial coarse object reconstruction is performed for multiple objects in general environments. Initial object segmentation using point cloud clustering and coarse segmentation is insensitive to parameters. Throughout this work the same parameters are used for all datasets. The result of this process is a coarse initial object segmentation and reconstruction for each object.
\begin{figure}
	\begin{center}
		\includegraphics[width = 0.99\linewidth]{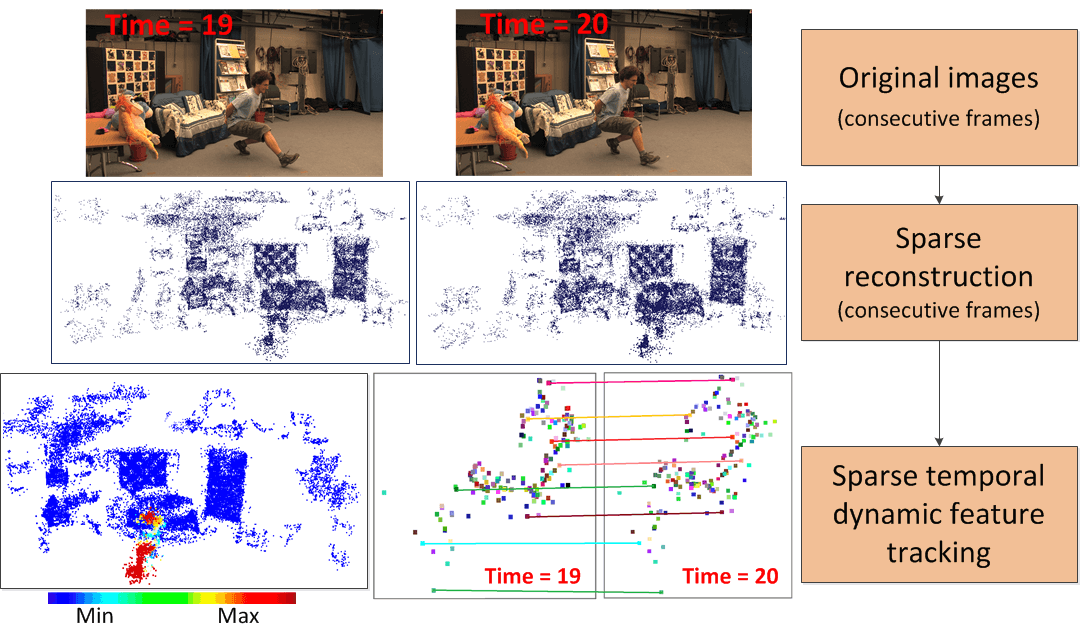}
		\caption{Sparse temporal dynamic feature tracking algorithm: Results on Odzemok dataset and Juggler dataset \cite{UnstructuredVBR10} captured with only moving cameras; Min and Max is the minimum and maximum movement in the 3D points respectively.}
		\label{fig:dynamicPts}
	\end{center}
\end{figure}
%
%
\subsection{Sparse-to-dense temporal reconstruction}
\label{sec:sparsecorrespondence}
Once the static scene reconstruction is obtained for the first frame, we perform temporally coherent reconstruction for dynamic objects at successive time instants instead of whole scene reconstruction for computational efficiency and to avoid redundancy.
Dynamic objects are identified from the temporal correspondence of sparse feature points (Section \ref{sec:sparsecorrespondence1}), shown in Figure \ref{fig:dynamicPts}. Sparse correspondence is used to propagate and obtain an initial model of the moving object for refinement (Section \ref{sec:sparsecorrespondence2}). The initial coarse reconstruction for each dynamic region is refined in the subsequent optimization step with respect to each camera view.
\begin{figure}
	\begin{center}
		\includegraphics[width = 0.99\linewidth]{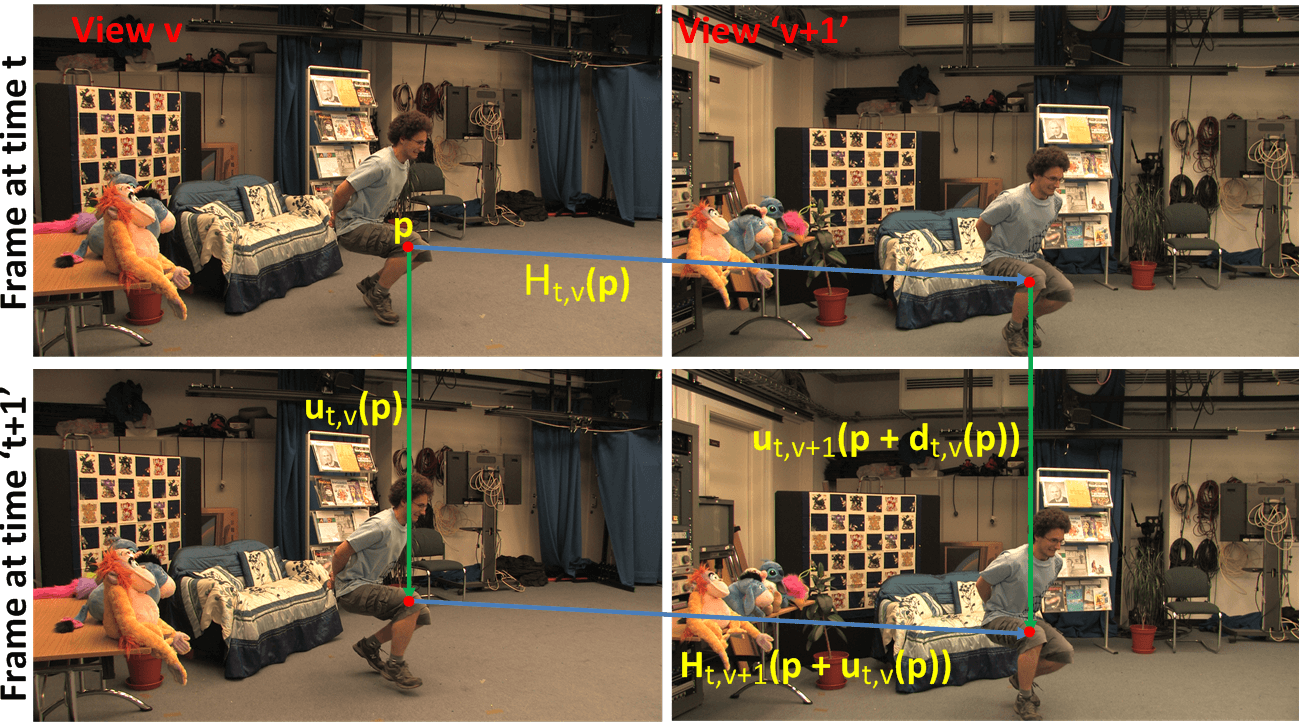}
		\caption{Spatio-temporal consistency check for 3D tracking for Odzemok dataset.}
		\label{fig:mvcc}
	\end{center}
\end{figure}
%
%
\subsubsection{Sparse temporal dynamic feature tracking} 
\label{sec:sparsecorrespondence1}
Numerous approaches have been proposed to track moving objects in 2D using either features or optical flow. However these methods may fail in the case of occlusion, movement parallel to the view direction, large motions and moving cameras.
To overcome these limitations we match the sparse 3D feature points obtained using SFD \cite{Mustafa15,Mustafa2019MSFDMS} from multiple wide-baseline views at each time instant.
The use of sparse 3D features is robust to large non-rigid motion, occlusions and camera movement. SFD detects sparse features which are stable across wide-baseline views and consecutive time instants for a moving camera and dynamic scene.  
Sparse 3D feature matches between consecutive time instants are back-projected to each view.
These features are matched temporally using SIFT descriptor to identify the corresponding moving points.
Robust matching is achieved by enforcing multiple view consistency for the temporal feature correspondence in each view as illustrated in Figure \ref{fig:mvcc}. 
Each match must satisfy the constraint:
\begin{eqnarray}
\left \| H_{t,v}(p) + u_{t,r}(p+H_{t,v}(p)) - u_{t,v}(p) 
-\right. \\ \nonumber
\left. H_{t,r}(p+u_{t,v}(p)) \right \|< \epsilon  
\end{eqnarray}
where $p$ is the feature image point in view $v$ at frame $t$, $H_{t,v}(p)$ is the disparity at frame $t$ from views $v$ and $r$, $u_{t,v}(p)$ is the temporal correspondence from frames $t$ to $t+1$ for view $v$.
The multi-view consistency check ensures that correspondences between any two views remain temporally consistent for  successive frames. 
Matches in the 2D domain are sensitive to camera movement and occlusion, hence we map the set of refined matches into 3D to make the system robust to camera motion. 
The Frobenius norm is applied on the 3D point gradients in all directions \cite{Zhang13} to obtain the `net' motion at each sparse point. The `net' motions between pairs of 3D points for consecutive time instants are ranked, and the top and bottom $5\%$ values are removed followed by Median filtering to identify the dynamic features. New objects are identified per frame from the clustered sparse reconstruction and are labelled as dynamic objects.
\begin{figure}[t]
	\begin{center}
		\includegraphics[width = 0.99\linewidth]{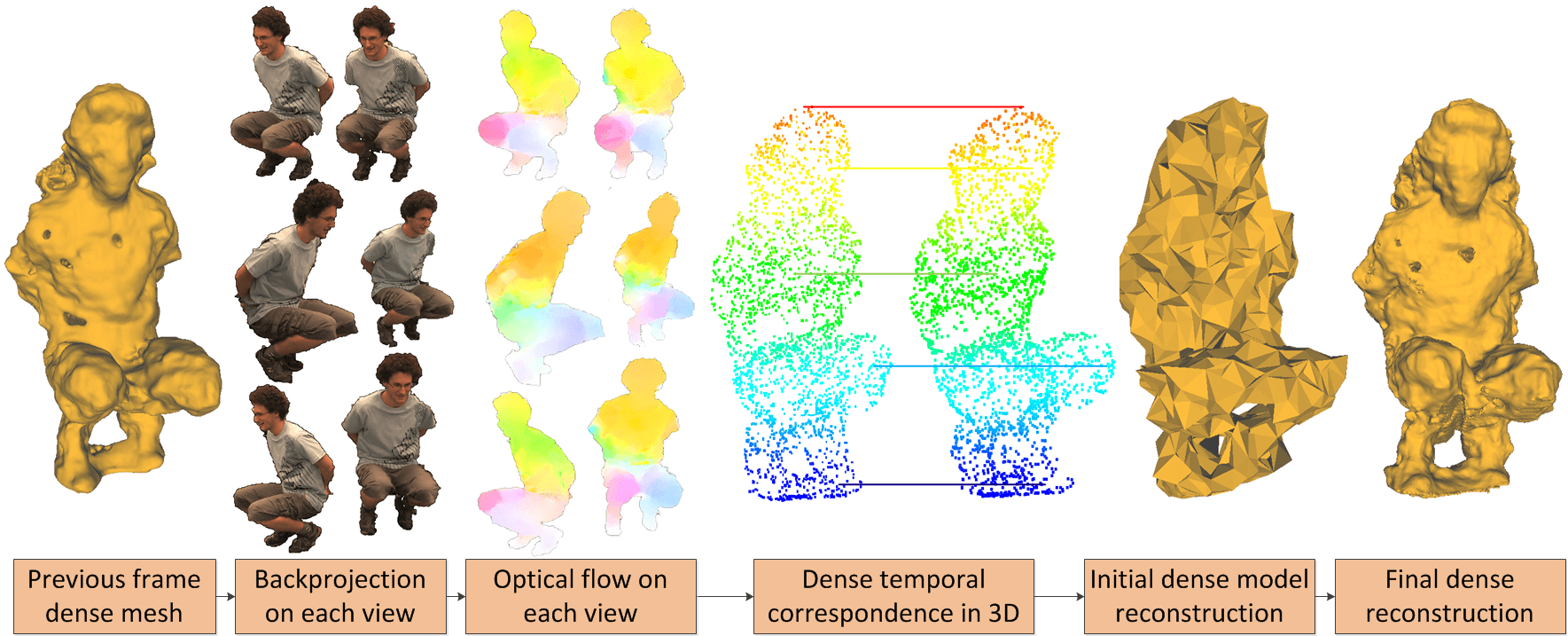}
		\caption{Initial sparse-to-dense model reconstruction workflow}
		\label{fig:initialmodel}
	\end{center}
\end{figure}
%
\vspace{-0.2cm}
\subsubsection{Sparse-to-dense model reconstruction} 
\label{sec:sparsecorrespondence2}
\vspace{-0.1cm}
Dynamic 3D feature points are used to initialize the segmentation and reconstruction of the initial model. 
This avoids the assumption of static backgrounds and prior scene segmentation commonly used to initialise multiple view reconstruction with a coarse visual-hull approximation \cite{Guillemaut2010}. 
Temporal coherence also provides a more accurate initialisation to overcome visual ambiguities at individual frames. 
Figure \ref{fig:initialmodel}  illustrates the use of temporal coherence for reconstruction initialisation and refinement.
Dynamic feature correspondence is used to identify the mesh for each dynamic object. This mesh is back projected on each view to obtain the region of interest. 
Lucas Kanade Optical flow \cite{Bouguet00} is performed on the projected mask for each view in the temporal domain using the dynamic feature correspondences over time as initialization. 
Dense multi-view wide-baseline correspondences from the previous frame are propagated to the current frame using the information from the flow vectors to obtain dense points in the current frame. 
The matches are triangulated in 3D to obtain a refined 3D dense model of the dynamic object for the current frame.\\
For dynamic scenes, to allow the introduction of new objects and object parts we use information from the cluster of sparse points for each dynamic object. The cluster corresponding to the dynamic features is identified and static points are removed. This ensures that the set of new points not only contain the dynamic features but also the unprocessed points which represent new parts of the object. These points are added to the refined sparse model of the dynamic object. To handle the new objects we detect new clusters at each time instant and consider them as dynamic regions. 
Once we have a set of dense 3D points for each dynamic object, Poisson surface reconstruction \cite{Kazhdan2006} is performed on the set of sparse points to obtain an initial coarse model of each dynamic region $\mathscr{R}$. The depth of region $\mathscr{R}$ is refined per view for each dynamic object at a per pixel level.

\begin{figure}[t]
	\begin{center}
		\includegraphics[width = 0.99\linewidth]{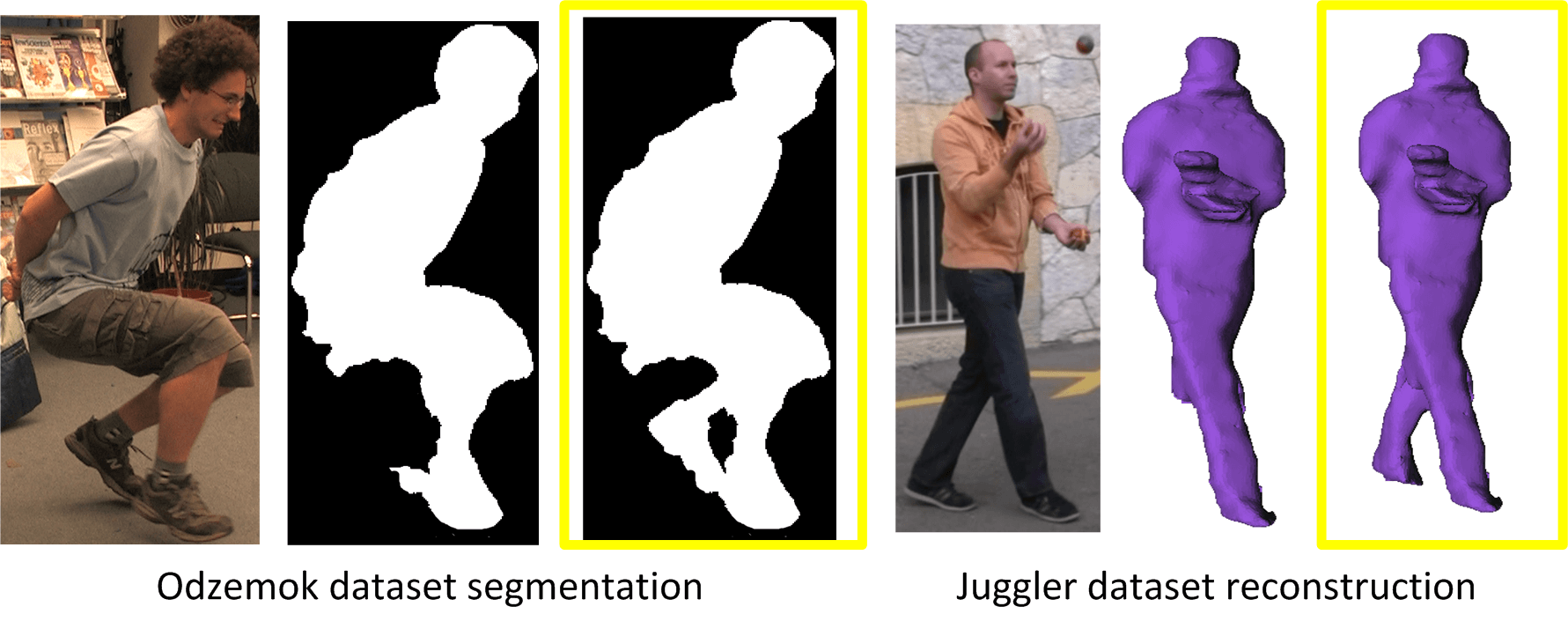}
		\caption{Improvement in segmentation for the Odzemok dataset and reconstruction for the Juggler dataset with temporal coherence (highlighted in yellow)}
		\label{fig:segTemporal}
	\end{center}
\end{figure}

The sparse-to-dense initial coarse reconstruction improves the quality of segmentation and reconstruction after the refinement. Examples of the improvement in segmentation and reconstruction for Odzemok \cite{cvssp3d} and Juggler \cite{UnstructuredVBR10} datasets are shown in Figure \ref{fig:segTemporal}. As observed limbs of the people is correctly reconstructed by using information from the previous frames in both the cases.
%

%
%
\subsection{Joint refinement of shape and segmentation}
\label{sec:4Doptimize}
The initial reconstruction $\mathscr{R}$ and segmentation ($\mathscr{R}$ projected in views) from dense temporal feature correspondence is refined using a joint optimization framework. View-dependent optimisation of depth is performed with respect to each camera which is robust to errors in camera calibration and initialisation. Calibration inaccuracies produce inconsistencies limiting the applicability of global reconstruction techniques which simultaneously consider all views; view-dependent techniques are more tolerant to such inaccuracies because they only use a subset of the views for reconstruction of depth from each camera view.

Our goal is to assign an accurate depth value from a set of depth values $\mathscr{D} = \left \{ d_{1},...,d_{\left|\mathscr{D} \right|-1} , \mathscr{U} \right \}$ and assign a layer label from a set of label values $\mathscr{L} = \left \{ l_{1},...,l_{\left|\mathscr{L} \right|} \right \}$ to each pixel $p$ for the region $\mathscr{R}$ of each dynamic object.
Each $d_{i}$ is obtained by sampling the optical ray from the camera and $\mathscr{U}$ is an unknown depth value to handle occlusions.
This is achieved by optimisation of a joint cost function \cite{Guillemaut2010} for label (segmentation) and depth (reconstruction): \\
$ E(l,d) = \lambda _{data}E_{data}(d) + \lambda _{contrast}E_{contrast}(l)  +$
\vspace{-0.2cm}
\begin{equation} \label{eq:costfunction}
\lambda _{smooth}E_{smooth}(l,d) + \lambda _{color}E_{color}(l) 
\vspace{-0.2cm}
\end{equation}
where, $d$ is the depth at each pixel, $l$ is the layer label for multiple objects and the cost function terms are defined in section \ref{sec:cost}. The equation consists of four terms: the data term is for the photo-consistency scores, the smoothness term is to avoid sudden peaks in depth and maintain the consistency and the color and contrast terms are to identify the object boundaries. Data and smoothness terms are common to solve reconstruction problems \cite{Bleyer11} and the color and contrast terms are used for segmentation \cite{Kolmogorov06}. This is solved subject to a geodesic star-convexity constraint on the labels $l$.
%
%
%
\begin{figure}
	\begin{center}
		\includegraphics[width = 0.99\linewidth]{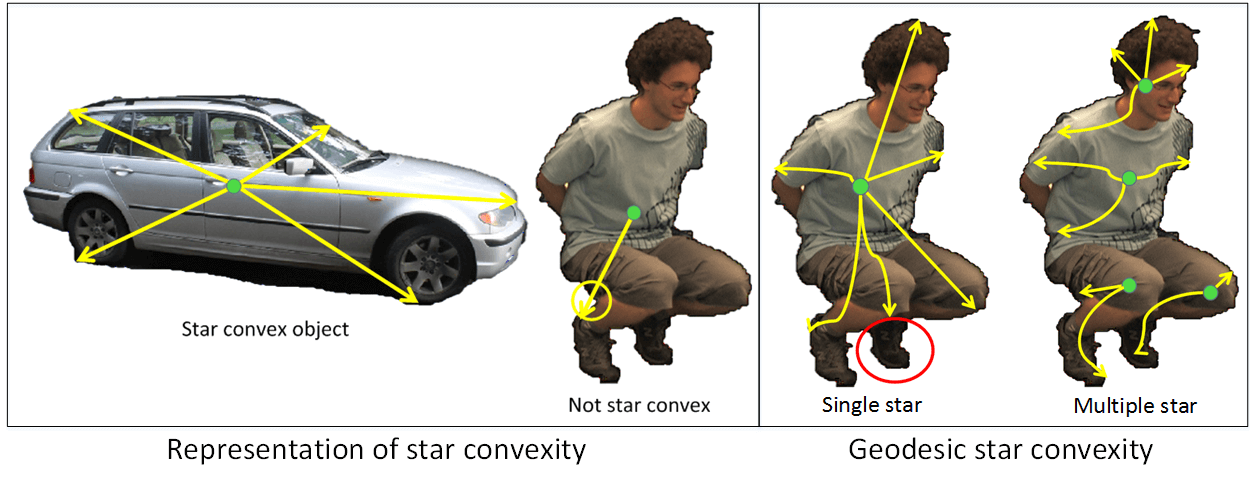}
		\caption{(a) Representation of star convexity:  The left object depicts example of star-convex objects, with a star center marked. The object on the right with a plausible star center shows deviations from star-convexity in the fine details, and (b) Multiple star semantics for joint refinement: Single star center based segmentation is depicted on the left and multiple star is shown on the right. }
		\label{fig:GSCconvex}
	\end{center}
\end{figure}
\subsubsection{Shape constraint for joint optimization}\label{sec:shape}
A novel shape constraint is introduced based on geodesic star convexity which has previously been shown to give improved performance in interactive image segmentation for structures with fine details (for example a person's fingers or hair)\cite{Gulshan10}. 
Previous methods used shape constraints by enforcing star convexity prior to improve segmentation \cite{Veksler08,Vicente08}. However star-convexity constraints fail for non-rigid objects (humans), as illustrated in Figure \ref{fig:GSCconvex}.
To handle complex objects the geodesic star convexity prior with multiple star centres was introduced in interactive segmentation for 2D objects \cite{Gulshan10}. The notion of connectivity was extended from Euclidean to geodesic space so that paths adapt to image data as opposed to straight Euclidean rays, thus extending visibility and reducing the number of star centers required. The union formed by multiple object seeds and geodesic paths form a geodesic forest \cite{Gulshan10}.

In this work the geodesic star-convexity shape constraint is automatically initialised for each view from the initial segmentation and sparse features to constrain the energy minimisation for joint multi-view reconstruction and segmentation. The shape constraint is based on the geodesic distance as star centres to which the object shape is restricted. This allows complex shapes to be segmented.
To automatically initialize the segmentation we use the sparse temporal feature correspondence as star centers (seeds) to build a geodesic forest.

\begin{figure}[t]
	\begin{center}
		\includegraphics[width = 0.99\linewidth]{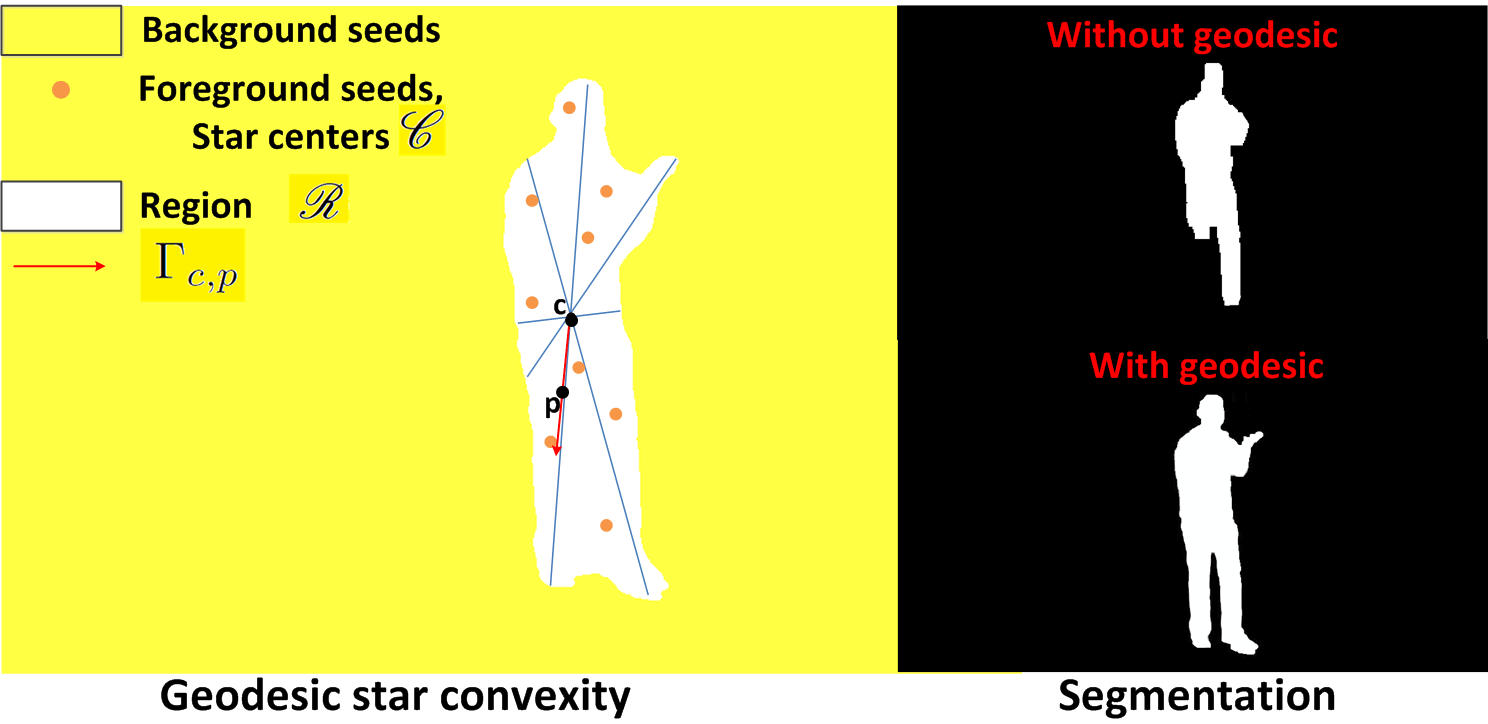}
		\caption{Geodesic star convexity: A region $\mathscr{R}$ with star centers $\mathscr{C}$ connected with geodesic distance $\Gamma _{c,p}$. Segmentation results with and without geodesic star convexity based optimization are shown on the right for the Juggler dataset.}
		\label{fig:geodesic}
	\end{center}
\end{figure}

The region outside the initial coarse reconstruction of all dynamic objects is initialized as the background seed for segmentation as shown in Figure \ref{fig:geodesic}. The shape of the dynamic object is restricted by this geodesic distance constraint that depends on the image gradient. 
Comparison with existing methods for multi-view segmentation demonstrates improvements in recovery of fine detail structure as illustrated in Figure \ref{fig:geodesic}.

In Equation \ref{eq:costfunction} a label $l$ is star convex with respect to center $c$, if every point $p\in l$ is visible to a star center $c$ via $l$ in the image $x$ which can be expressed as an energy cost:
\begin{equation} \label{eq:singleStar}
E^{\star}(l \rvert x, c) = \sum_{p\in {R}} \sum_{q \in \Gamma _{c,p}} E_{p,q}^{\star}(l_p , l_q)
\end{equation}
\begin{equation} \label{eq:geodesic2}
\forall q \in \Gamma_{c,p} , \text{      } E_{p,q}^{\star} = \left\{\begin{matrix}
\infty \text{          if  } l_p \neq l_q\\ 
0      \text{                      otherwise              }\\ 
\end{matrix}\right.
\end{equation}
where $\forall p \in {R}: p \in l \Leftrightarrow l_p = 1 $ and $\Gamma _{c,p}$ is the geodesic path joining $p$ to the star center $c$ given by:
\begin{equation} \label{eq:geodesic3}
\Gamma_{c,p} = \argmin_{\Gamma \in {P}_{c,p}} {L}(\Gamma)
\end{equation}
where ${P}_{c,p}$ denotes the set of all discrete paths between $c$ and $p$ and ${L}(\Gamma)$ is the length of discrete geodesic path as defined in \cite{Gulshan10}.  
In the case of image segmentation the gradients in the underlying image provide information to compute the discrete paths between each pixel and star centers and ${L}(\Gamma)$ is defined below:
\begin{equation}\nonumber
{L}(\Gamma) = \sum_{i = 1}^{N_{D} - 1}\sqrt{  (1 - \delta _{g})j(\Gamma^{i},\Gamma^{i+1})^{2} +   \delta _{g}\left \|  \bigtriangledown I (\Gamma^{i})  \right \|^{2} }
\end{equation}
where $\Gamma$ is an arbitrary parametrized discrete path with $N_{D}$ pixels given by $\left \{ \Gamma ^{1} , \Gamma ^{2}, \cdots \Gamma ^{N_D} \right \}$, $j(\Gamma^{i},\Gamma^{i+1})$ is the Euclidean distance between successive pixels, and the quantity $\left \|  \bigtriangledown I (\Gamma^{i})  \right \|^{2}$ is a finite difference approximation of the image gradient between the points $\left (  \Gamma^{i}, \Gamma^{i+1}\right )$. The parameter weights $\delta _{g}$ the Euclidean distance with the geodesic length. Using the above definition, one can define the geodesic distance as defined in Equation \ref{eq:geodesic3}.  
%

%
\begin{figure}
	\begin{center}
		\includegraphics[width = 0.99\linewidth]{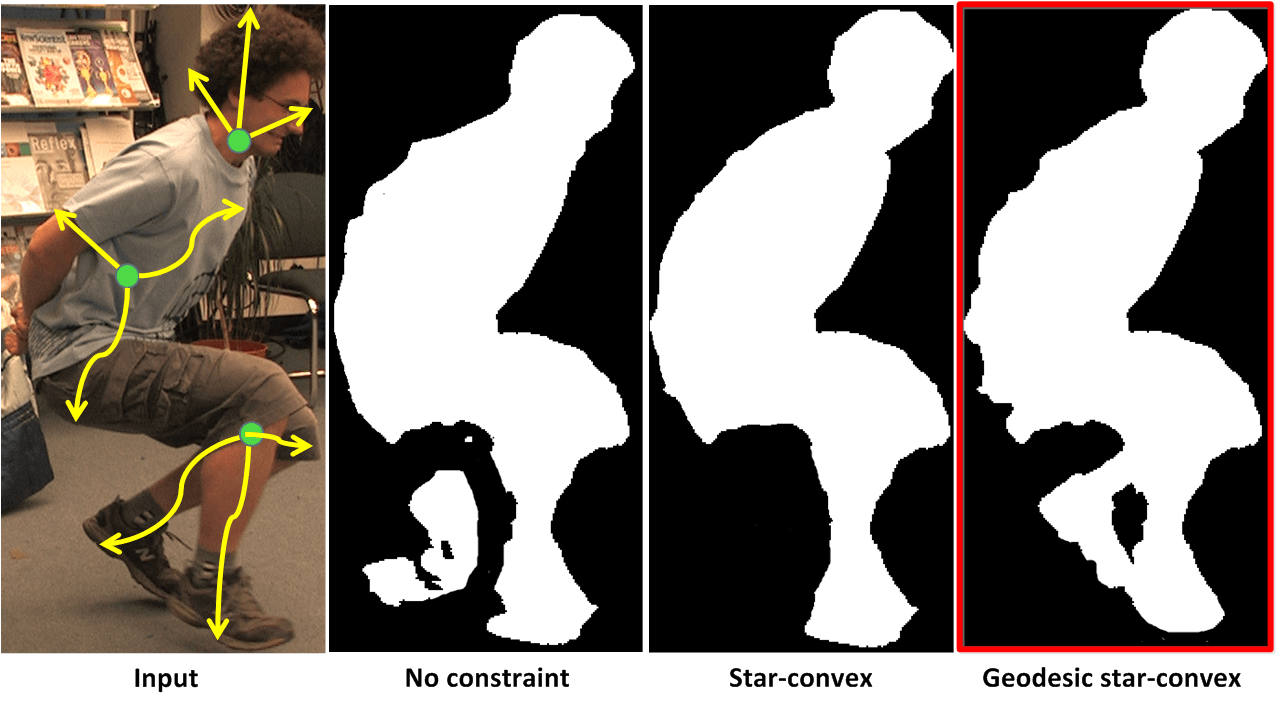}
		\caption{Segmentation comparison results with no constraint, star convexity constraint and geodesic star convexity constraint for Odzemok dataset. }
		\label{fig:segGSC}
	\end{center}
\end{figure}
%
An extension of single star-convexity is to use multiple stars to define a more general class of shapes. Introduction of multiple star centers reduces the path lengths and increases the visibility of small parts of objects like small limbs as shown in Figure \ref{fig:GSCconvex}.
Hence Equation \ref{eq:singleStar} is extended to multiple stars. A label $l$ is star convex with respect to center $c_{i}$, if every point $p\in l$ is visible to a star center  $c_{i}$ in set $\mathscr{C} = \left \{ c_{1},...,c_{N_T}  \right \}$ via $l$ in the image $x$, where $N_T$ is the number of star centers \cite{Gulshan10}. This is expressed as an energy cost:
\begin{equation} \label{eq:geodesic1}
E^{\star}(l \rvert x, \mathscr{C}) = \sum_{p\in {R}} \sum_{q \in \Gamma _{c,p}} E_{p,q}^{\star}(l_p , l_q)
\end{equation}
In our case all the correct temporal sparse feature correspondences are used as star centers, hence the segmentation will include all the points which are visible to these sparse features via geodesic distances in the region $\mathscr{R}$, thereby employing the shape constraint. Since the star centers are selected automatically, the method is unsupervised. Comparison of segmentation constraint with geodesic multi-star convexity against no constraints and Euclidean multi-star convexity constraint is shown in Figure \ref{fig:segGSC}. The figure demonstrates the usefulness of the proposed approach with an improvement in segmentation quality on non-rigid complex objects.
The energy in the Equation \ref{eq:costfunction} is minimized as follows:
\begin{equation} \label{eq:minimize}
\underset{s.t.}{min_{(l,d)}} \text{         }\underset{l\epsilon S^{\star }(\mathscr{C})}{E(l,d)}   \Leftrightarrow \min_{(l,d)}E(l,d) + 	E^{\star}(l \rvert x, \mathscr{C})
\end{equation}
where $ S^{\star }(\mathscr{C})$ is the set of all shapes which lie within the geodesic distances with respect to the centers in $\mathscr{C}$.
Optimization of Equation \ref{eq:minimize}, subject to each pixel $p$ in the region $\mathscr{R}$ being at a geodesic distance $\Gamma_{c,p}$ from the star centers in the set $\mathscr{C}$, is performed using the $\alpha$-expansion algorithm for a pixel $p$ by iterating through the set of labels in $\mathscr{L} \times \mathscr{D}$ \cite{Boykov01}. Graph-cut is used to obtain a local optimum \cite{Kolmogorov04}. 
%
%
\begin{figure}
	\begin{center}
		\includegraphics[width = 0.99\linewidth]{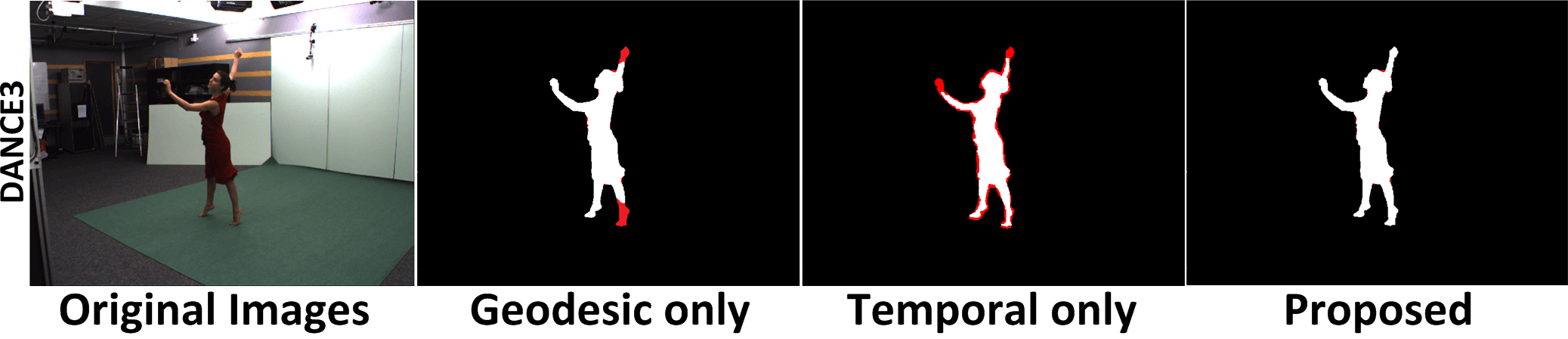}
		\caption{Comparison of segmentation with introduction of temporal coherence, Geodesic star convexity(GSC) and proposed method (GSC and temporal coherence) for Dance2 dataset.}
		\label{fig:proposed}
	\end{center}
\end{figure}
The improvements in the results using geodesic star convexity in the framework is shown in Figure \ref{fig:geodesic} and by using temporal coherence is shown in Figure \ref{fig:segTemporal}. Figure \ref{fig:proposed} shows improvements using geodesic shape constraint, temporal coherence and combined proposed approach for Dance2 \cite{4DInria} dataset.
%
\begin{figure}
	\begin{center}
		\includegraphics[width = 0.99\linewidth]{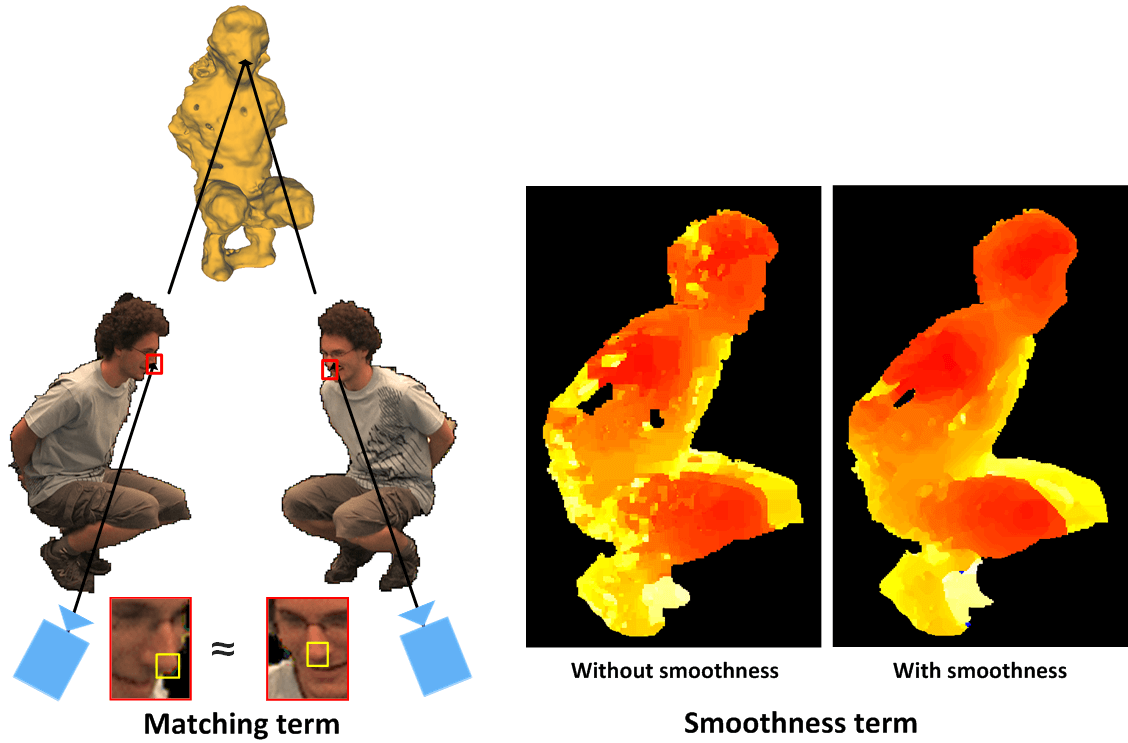}
		\caption{Illustration of matching and smoothness term for the energy minimization}
		\label{fig:colorNsmooth}
	\end{center}
\end{figure}
%
%
%
%
\vspace{-0.4cm}
\subsubsection{Energy cost function for joint optimization}\label{sec:cost}
\vspace{-0.1cm}
For completeness in this section we define each of the terms in Equation \ref{eq:costfunction}, these are based on previous terms used for joint optimisation over depth for each pixel introduced in \cite{MustafaICCV15}, with modification of the color matching term to improve robustness and extension to multiple labels. \\
\textbf{Matching term:}
The data term for matching between views is specified as a measure of photo-consistency (Figure \ref{fig:colorNsmooth}) as follows:\\
$ E_{data}(d) = \sum_{p\in \mathscr{P}} e_{data}(p, d_{p}) = $
\begin{equation} \label{eq:matching1}
\begin{cases}
M(p, q)  = \sum_{i \in \mathscr{O}_{k}}m(p,q) ,& \text{if } d_{p}\neq \mathscr{U}\\
M_{\mathscr{U}}, & \text{if } d_{p} = \mathscr{U}\\
\end{cases}
\vspace{-0.5cm}
\end{equation}
where $\mathscr{P}$ is the 4-connected neighbourhood of pixel $p$, $M_{\mathscr{U}}$ is the fixed cost of labelling a pixel unknown and $q$ denotes the projection of the hypothesised point $P$ in an auxiliary camera where $P$ is a $3D$ point along the optical ray passing through pixel $p$ located at a distance $d_{p}$ from the reference camera. $\mathscr{O}_{k}$ is the set of $k$ most photo-consistent pairs.\\
For textured scenes Normalized Cross Correlation (NCC) over a squared window is a common choice \cite{Seitz06}. The NCC values range from -1 to 1 which are then mapped to non-negative values by using the function $1 - NCC$. A maximum likelihood measure \cite{Larry92} is used in this function for confidence value calculation between the center pixel $p$ and the other pixels $q$ and is based on the survey on confidence measures for stereo \cite{Hu12}. The measure is defined as:
\begin{equation} \label{eq:matching4}
m(p,q) = \frac{exp\tfrac{c_{min}}{2\sigma_{i}^{2}}}{\sum_{(p,q) \in \mathscr{N}} exp\tfrac{-(1-NCC(p,q))}{2\sigma_{i} ^{2}}}
\vspace{-0.2cm}
\end{equation}
where $\sigma_{i} ^{2}$ is the noise variance for each auxiliary camera $i$; this parameter was fixed to $0.3$. $\mathscr{N}$ denotes the set of interacting pixels in $\mathscr{P}$. $c_{min}$ is the minimum cost for a pixel obtained by evaluating the function $(1 - NCC(.,.))$ on a $15 \times 15$ window.\\
%
%
\textbf{Contrast term:}
Segmentation boundaries in images tend to align with contours of high contrast and it is desirable to represent this as a constraint in stereo matching. A consistent interpretation of segmentation-prior and contrast-likelihood is used from \cite{Kolmogorov06}. We used a modified version of this interpretation in our formulation to preserve the edges by using Bilateral filtering \cite{Tomasi1998} instead of Gaussian filtering. The contrast term is as follows:
\vspace{-0.2cm}
\begin{equation} \label{eq:contrast1}
E_{contrast}(l) =  \sum_{p,q \in \mathscr{N}} e_{contrast}(p,q,l_p,l_q)
\end{equation}
\vspace{-0.3cm}
\begin{equation} \label{eq:contrast2} \nonumber
e_{contrast}(p,q,l_p,l_q)=  
\begin{cases}
0, & \text{if } (l_{p} =  l_{q})\\
\frac{1}{1+\epsilon }( \epsilon + exp^{-C(p,q)}),  & \text{otherwise}
\end{cases}
\end{equation}
$\left \| \cdot  \right \|$ is the $L_{2}$ norm and $\epsilon = 1$. The simplest choice for $C(p,q)$ would be the squared Euclidean color distance between intensities at pixel $p$ and $q$ as used in \cite{Guillemaut2010}. We propose a term for better segmentation as  $C(p,q) = \frac{\left \| B(p) - B(q) \right \|^{2}}{2 \sigma _{pq}^{2} d_{pq}^{2} }$ where $B(.)$ represents the bilateral filter, $d_{pq}$ is the Euclidean distance between $p$ and $q$, and $\sigma _{pq} = \left \langle\frac{\left \| B(p) - B(p)\right \|^{2}}{d_{pq}^{2}}\right\rangle$
This term enables to remove the regions with low photo-consistency scores and weak edges and thereby helps in estimating the object boundaries. \\
\textbf{Smoothness term:}
This term is inspired by \cite{Guillemaut2010} and it ensures the depth labels vary smoothly within the object reducing noise and peaks in the reconstructed surface. This is useful when the photo-consistency score is low and insufficient to assign depth to a pixel (Figure \ref{fig:colorNsmooth}). It is defined as:
\vspace{-0.2cm}
\begin{equation} \label{eq:smooth}
E_{smooth}(l,d) = \sum_{(p,q)\in \mathscr{N}} e_{smooth}(l_p,d_{p},l_q,d_{q}) =
\vspace{-0.2cm}
\end{equation}
\vspace{-0.5cm}
\begin{equation} \nonumber
\begin{cases}
min(\left | d_{p} - d_{q} \right |, d_{max}),& \text{if } l_{p} =  l_{q} \text{ and }  d_{p},d_{q}\neq \mathscr{U}\\
0,              & \text{if } l_{p} =  l_{q} \text{ and } d_{p},d_{q} = \mathscr{U}\\
d_{max},  & \text{otherwise}
\vspace{-0.2cm}
\end{cases}
\end{equation}
$d_{max}$ is set to 50 times the size of the depth sampling step for all datasets.\\
\textbf{Color term:}
This term is computed using the negative log likelihood \cite{Kolmogorov04} of the color models learned from the foreground and background markers. The star centers obtained from the sparse 3D features are foreground markers and for background markers we consider the region outside the projected initial coarse reconstruction for each view. The color models use GMMs with 5 components each for Foreground/Background mixed with uniform color models \cite{Das09} as the markers are sparse. 
\vspace{-0.1cm}
\begin{equation} \label{eq:color}
E_{color}(l) = \sum_{p\in \mathscr{P}} -log P(I_{p}\rvert l_{p})
\vspace{-0.5cm}
\end{equation}
where $P(I_{p}\rvert l_{p} = l_i)$ denotes the probability at pixel $p$ in the reference image belonging to layer $l_i$.
%
%
\begin{table}
	\centering
	\setlength{\tabcolsep}{3pt}
	\scalebox{0.95}{
		\begin{tabular}{|l|c|c|c|c|c|c|}
			\hline
			\textbf{Datasets} &  \textbf{Resolution} & \textbf{No. of views} & \textbf{Baseline} &  \textbf{Type} \\ \hline
			Office\cite{cvssp3d} & $1920 \times 1080$ & 8(all S) & $25\degree$-$35\degree$ & dynamic \\
			Juggler\cite{UnstructuredVBR10} & $960 \times 544$ & 6(all M) & $15\degree$-$45\degree$ & dynamic \\
			Dance1\cite{cvssp3d} & $1920 \times 1080$ & 8(all S) & $20\degree$-$30\degree$ & dynamic \\
			Odzemok\cite{cvssp3d} & $1920 \times 1080$ & 8(2 M) & $15\degree$-$30\degree$ & dynamic \\
			Dance2\cite{4DInria} & $780 \times 582$ & 8(all S) & $35\degree$-$45\degree$ & dynamic \\
			Magician\cite{UnstructuredVBR10}& $960 \times 544$ & 6(all M) & $15\degree$-$45\degree$ & dynamic \\
			Couch\cite{Kowdle12} & $640 \times 480$ & 9(all S) & $25\degree$-$30\degree$ & static \\
			Chair\cite{Kowdle12} & $640 \times 480$ & 17(all S) & $5\degree$-$8\degree$ & static \\
			Car\cite{Kowdle12} & $640 \times 480$ & 16(all S) & $5\degree$-$8\degree$ & static \\
			\hline
	\end{tabular}}
	\caption{Properties of all datasets where \textbf{Type} represents whether the data is static or dynamic. In \textbf{No. of views} S stands for static cameras and M for moving cameras.}
	\label{t_dataset}
\end{table}
%
\begin{figure}[]
	\begin{center}
		\includegraphics[width =0.99\linewidth]{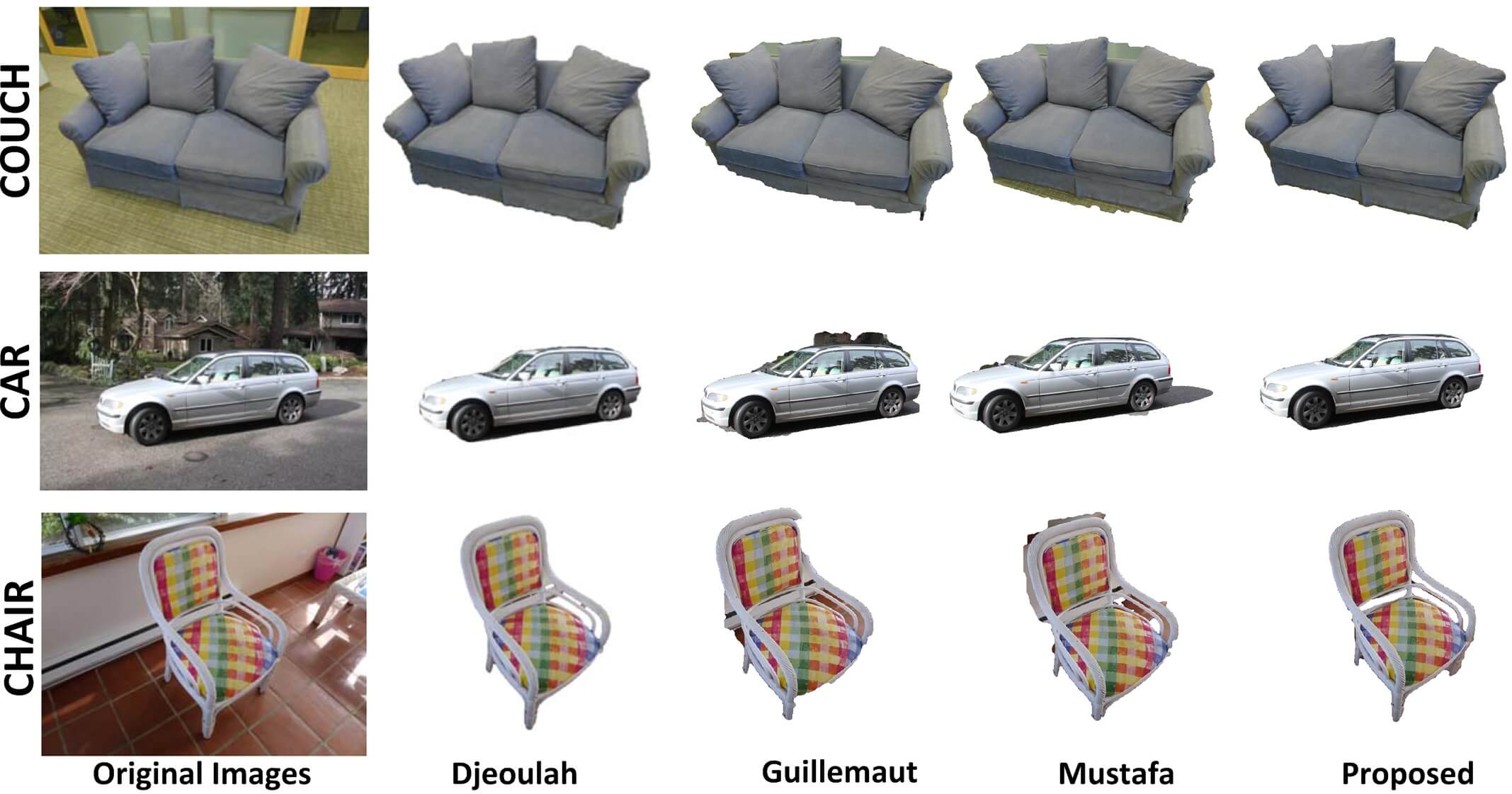}
		\caption{Comparison of segmentation on benchmark static datasets using geodesic star-convexity.}
		\label{fig:staticSeg}
	\end{center}
\end{figure}
\vspace{-0.4cm}
\section{Results and Performance Evaluation} 
\label{sec:results}
\vspace{-0.1cm}
The proposed system is tested on publicly available multi-view research datasets of indoor and outdoor scenes, details of datasets explained in Table \ref{t_dataset}. The parameters used for all the datasets are defined in Table \ref{parameters}. More information is available on the website\footnotemark \footnotetext{http://cvssp.org/projects/4d/4DRecon/}.
%
\begin{table}
	\setlength{\tabcolsep}{4pt}
	\scalebox{1}{
		\begin{tabular}{|l|c|c|c|c|}
			\hline
			& $\lambda_{data}$ & $\lambda_{c}$ & $\lambda_{smooth}$ & $\lambda_{color}$ \\ \hline
			{\small Magician/Dance2} & 0.4 & 5.0 & .0005 & 0.6 \\ 
			Juggler & 0.5 & 5.0 & .0005 & 0.4 \\
			{\small Odzemok/Dance1/Office} & 0.4 & 3.0 & .001 & 0.6 \\ \hline
	\end{tabular}}
	\caption{Parameters used for all datasets: $\lambda_{c}$ represents $\lambda_{contrast}$}
	\label{parameters}
\end{table}
%
%
\begin{table}
	\begin{center}
		\setlength{\tabcolsep}{2pt}
		\scalebox{0.925}{
			\begin{tabular}{|l|c|c|c|c|c|c|c|} 
				\hline
				Dataset & Kowdle & Djelouah & Guillemaut & Mustafa & Proposed \\
				\hline
				Couch  & $99.6\pm0.1$ & $99.0\pm0.2$ & $97.0\pm0.3$ & $98.5\pm0.2$ & \textbf{99.7$\pm$0.3}\\
				Chair  & \textbf{99.2$\pm$0.4} & $98.6 \pm 0.3$ & $97.9 \pm 0.5$  & $98.0 \pm 0.5$ & $99.1 \pm 0.3$ \\
				Car    & $98.0 \pm 0.7$ & $97.0 \pm 0.8$ & $95.0 \pm 0.7$ & $97.6 \pm 0.3$ & \textbf{98.6$\pm$0.4} \\
				\hline
		\end{tabular}}
	\end{center}
	\vspace{-0.2cm}
	\caption{Static segmentation completeness comparison with existing methods on benchmark datasets ($\%$)}
	\label{staticSegResults}
\end{table}
%
\begin{figure}
	\begin{center}
		\includegraphics[width=0.99\linewidth]{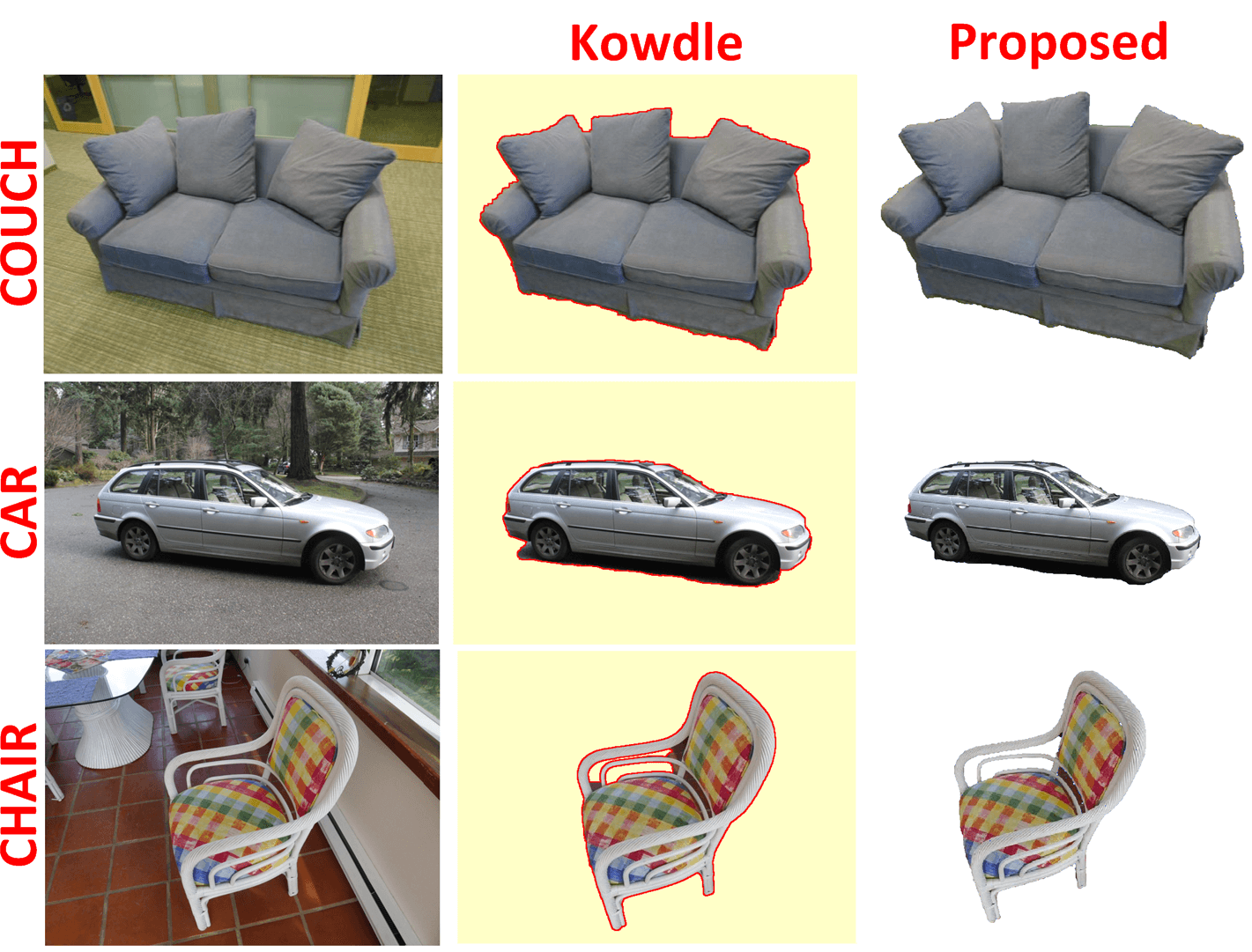}
	\end{center}
	\vspace{-0.2cm}
	\caption{Comparison of segmentation with Kowdle \cite{Kowdle12}.}
	\vspace{0.1cm}
	\label{fig:kowdle}
\end{figure}
\begin{figure}
	\begin{center}
		\includegraphics[width =0.99\linewidth]{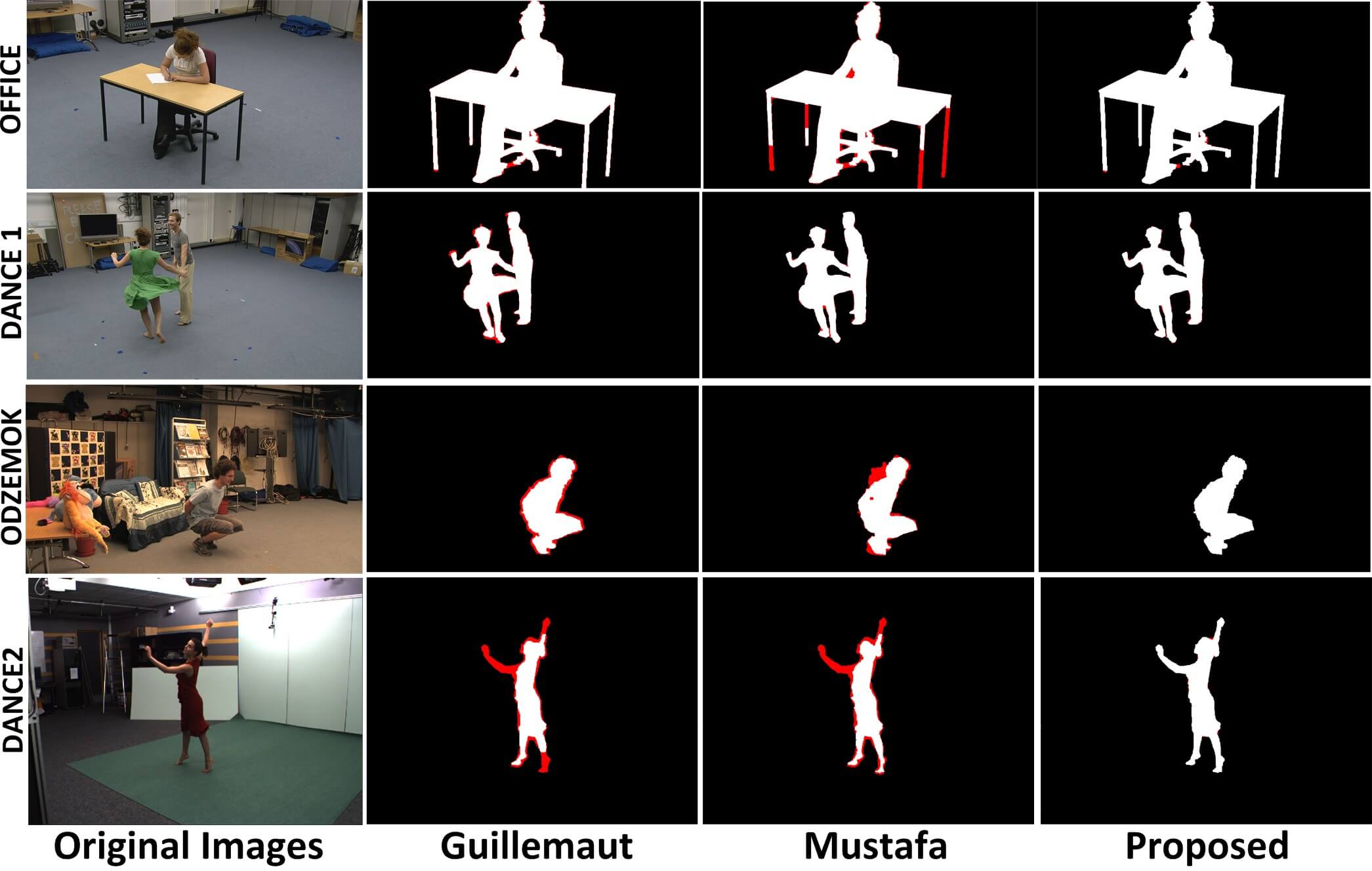}
		\caption{Segmentation results for dynamic scenes  (Error against ground-truth is highlighted in red).}
		\label{fig:segment}
	\end{center}
\end{figure}
\begin{figure*}
	\begin{center}
		\includegraphics[width =0.99\linewidth]{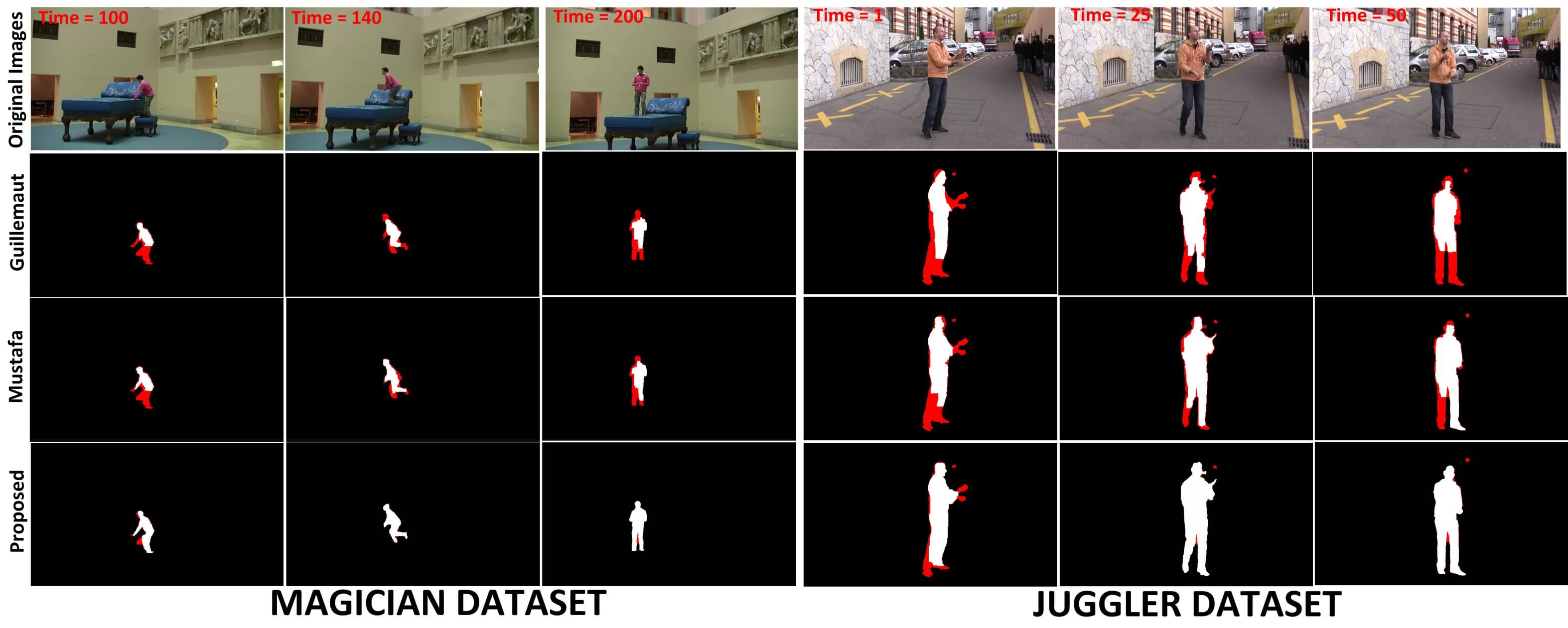}
		\caption{Segmentation results for dynamic scenes on sequence of frames (Error against ground-truth is highlighted in red).}
		\vspace{-0.7cm}
		\label{fig:dynamicSeg}
	\end{center}
\end{figure*}
%
\begin{figure}
	\begin{center}
		\includegraphics[width = 0.99\linewidth]{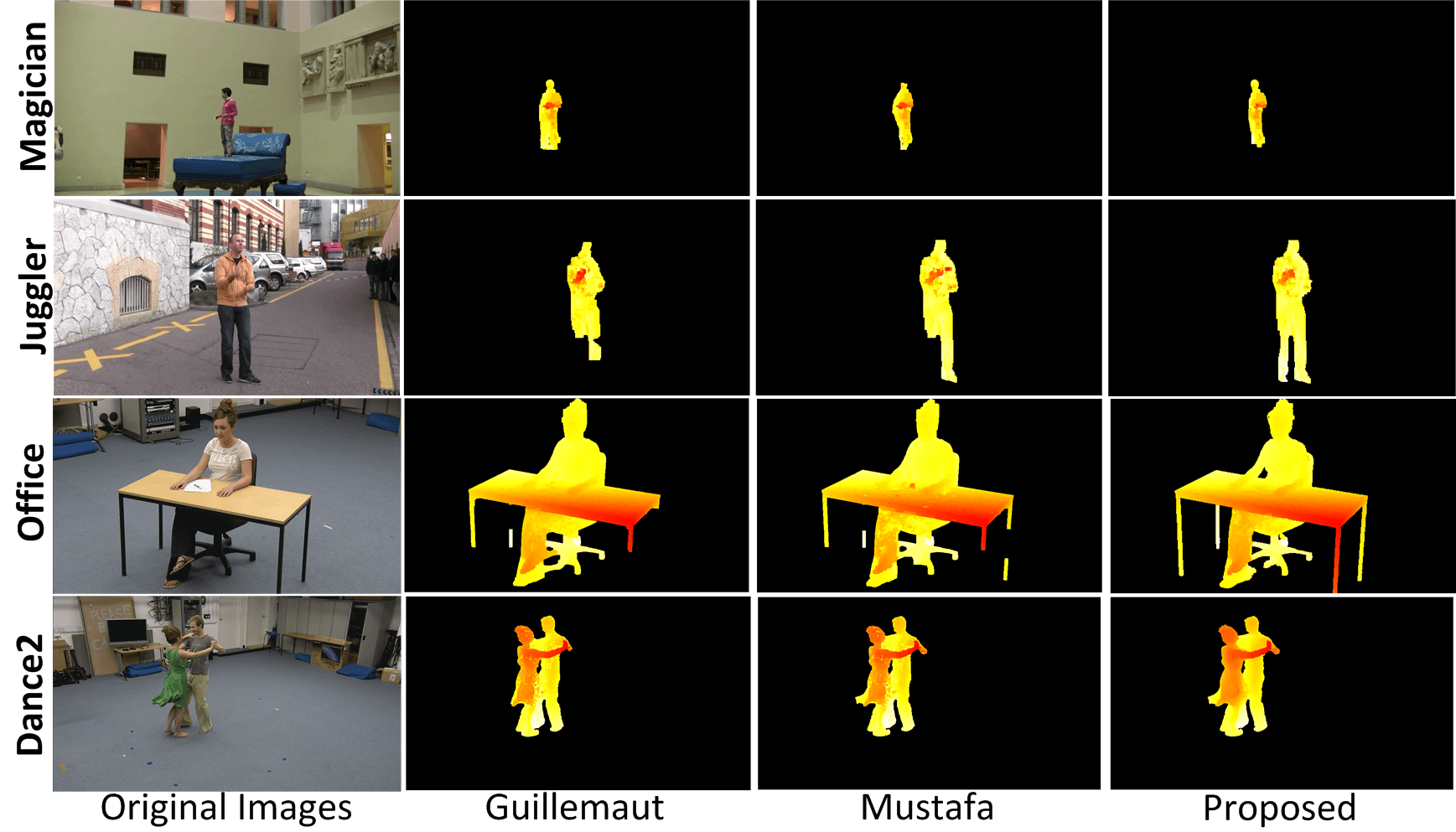}
		\vspace{-0.35cm}
		\caption{Comparison of depth maps against existing methods for two  indoor and two outdoor benchmark datasets.}
		\label{fig:depthMagnJug}
	\end{center}
\end{figure}
%
%
%
\begin{table}[h]
	\setlength{\tabcolsep}{3pt}
	\scalebox{0.89}{
		{\begin{tabular}{|l|c|c|c|c|c|c|c|}
			\hline
			& Magician & Juggler & Odzemok & Dance1 & Office & Dance2 \\
			\hline
			Guillemaut & $68.0$ & $84.6 $ & $90.1$ & $99.2$ & $99.3$ & $98.6$ \\
			Mustafa & $88.7$ & $87.9 $ & $89.9$ & $99.4$ & $99.0$ & $99.0$ \\
			$P_{contrast}$ & $79.6$ & $88.9$ & $90.5$ & $99.1$ & $99.2$ & $98.8$ \\
			$P_{smooth}$ & $89.9$ & $90.6 $ & $90.8$ & $99.2$ & $99.2$ & $98.9$  \\
			$P_{ds}$ & $78.3$ & $86.2$ & $90.0$ & $99.3$ & $99.1$ & $98.7$  \\
			Proposed & \textbf{91.2} & \textbf{93.3} & \textbf{91.8} & \textbf{99.5} & \textbf{99.4} & \textbf{99.0} \\
			\hline
	\end{tabular}}}
	\caption{Dynamic scene segmentation completeness ($\%$): $P_{smooth}=Proposed-E_{smooth}$, $P_{contrast}=Proposed-E_{contrast}$, $P_{ds}=Proposed- E_{data}-E_{smooth}$ }
	\label{dynamicSegResults}
\end{table}
\vspace{-0.25cm}
\subsection{Multi-view segmentation evaluation}
\label{sec:segresults}
Segmentation is evaluated against the state-of-the-art methods for multi-view segmentation Kowdle \cite{Kowdle12} and Djelouah \cite{Djelouah13} for static scenes and joint segmentation reconstruction methods Mustafa \cite{MustafaICCV15} (per frame) and Guillemaut \cite{Guillemaut3dv} (using temporal information) for both static and dynamic scenes.\\
For static multi-view data the segmentation is initialised as detailed in Section \ref{sec:method} followed by refinement using the constrained optimisation Section \ref{sec:4Doptimize}. For dynamic scenes the full pipeline with temporal coherence is used as detailed in \ref{sec:method}.
Ground-truth is obtained by manually labelling the foreground for Office, Dance1 and Odzemok dataset, and for other datasets ground-truth is available online.
We initialize all approaches by the same proposed initial coarse reconstruction for fair comparison.
To evaluate the segmentation we measure completeness as the ratio of intersection to union with ground-truth \cite{Kowdle12}. Comparisons are shown in Table \ref{staticSegResults} and Figure \ref{fig:staticSeg}, \ref{fig:kowdle} for static benchmark datasets. Comparison for dynamic scene segmentations are shown in Table \ref{dynamicSegResults} and Figure \ref{fig:segment}, \ref{fig:dynamicSeg}. 
Results for multi-view segmentation of static scenes are more accurate than Djelouah, Mustafa, and Guillemaut, and comparable to Kowdle with improved segmentation of some detail such as the back of the chair. 
We also perform ablative analysis on Equation \ref{eq:costfunction} by removing $E_{data}$, $E_{smooth}$ and $E_{contrast}$ terms. Results demonstrate that joint depth ($E_{data}$, $E_{smooth}$) and segmentation estimation improves the result and contrast information ($E_{contrast}$) helps in improving the quality of the segmentation.

For dynamic scenes the geodesic star convexity based optimization  together with temporal consistency gives improved segmentation of fine detail such as the legs of the table in the Office dataset and limbs of the person in the Juggler, Magician and Dance2 datasets in Figure \ref{fig:segment} and \ref{fig:dynamicSeg}. This overcomes limitations of previous multi-view per-frame segmentation. 
%
%
\begin{table}
	\begin{center}
		\begin{tabular}{|l|c|c|c|c|} 
			\hline
			Dataset & Furukawa & Guillemaut & Mustafa & Ours \\
			\hline
			Dance1   & 326  & 493  & 295  &  \textbf{254 } \\
			Magician & \textbf{311} & 608 & 377  &  325  \\
			Odzemok  & 381  & 598  & 394  &  \textbf{363 } \\
			Office   & 339  & 533  & 347  &  \textbf{291 } \\
			Juggler  & 394  & 634  & 411  &  \textbf{378 } \\
			Dance2   & 312  & 432  & 323  &  \textbf{278 } \\
			\hline
		\end{tabular}
	\end{center}
	\vspace{-0.35cm}
	\caption{Comparison of computational efficiency for dynamic datasets (time in seconds (s))}
	\label{time}
\end{table}
%
%
%
\begin{figure*}
	\begin{center}
		\includegraphics[width =0.99\linewidth]{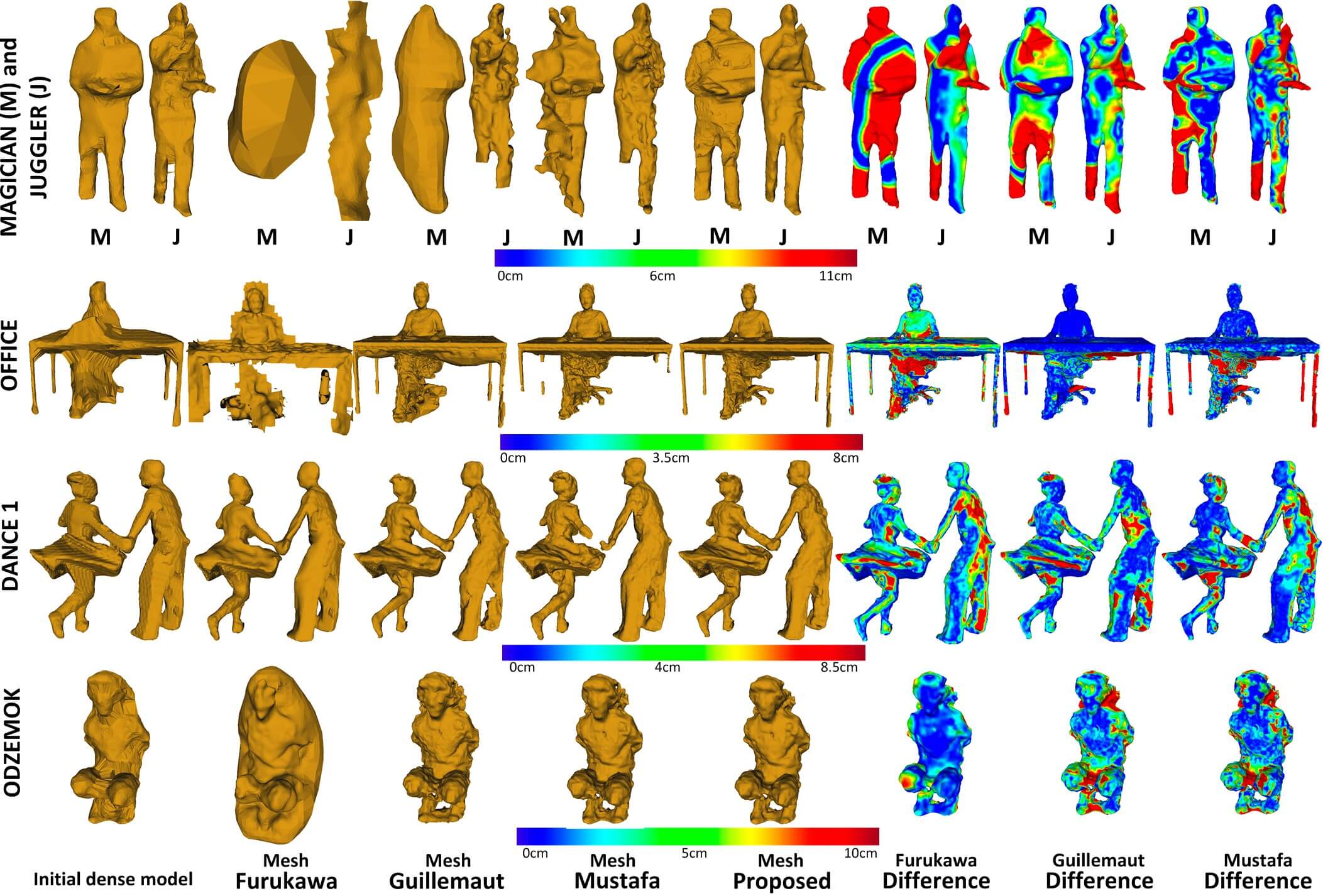}
		\caption{Reconstruction result mesh comparison against state-of-the-art methods with errors shown in the last three columns}
		\vspace{-0.7cm}
		\label{fig:meshes}
	\end{center}
\end{figure*}
%
%
\begin{figure*}
	\begin{center}
		\includegraphics[width=0.99\linewidth]{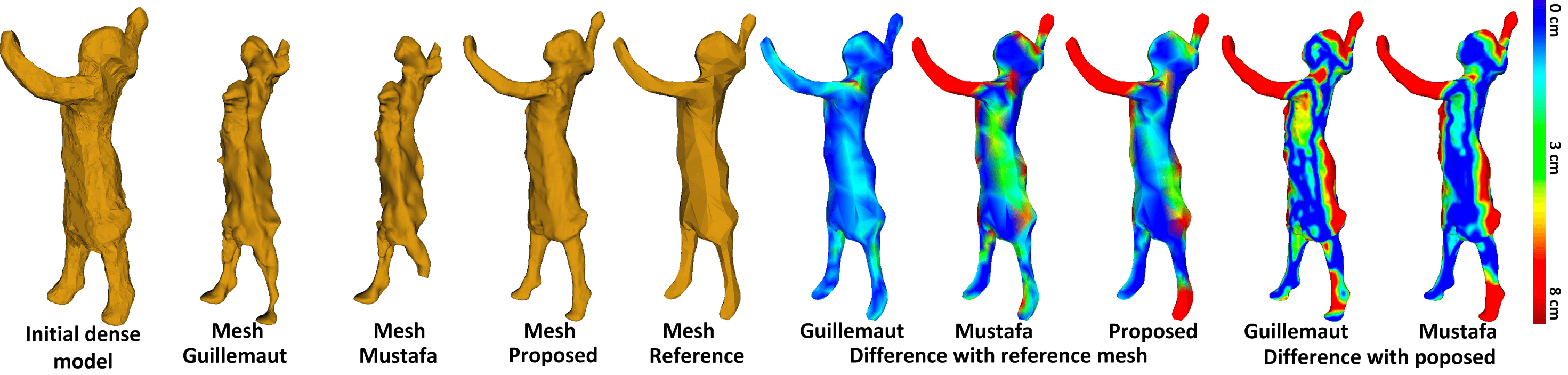}
		\caption{Reconstruction result comparison with reference mesh and proposed for Dance2 benchmark dataset}
		\vspace{-0.7cm}
		\label{fig:dance2}
	\end{center}
\end{figure*}
%
%
\begin{figure}
	\begin{center}
		\includegraphics[width=0.99\linewidth]{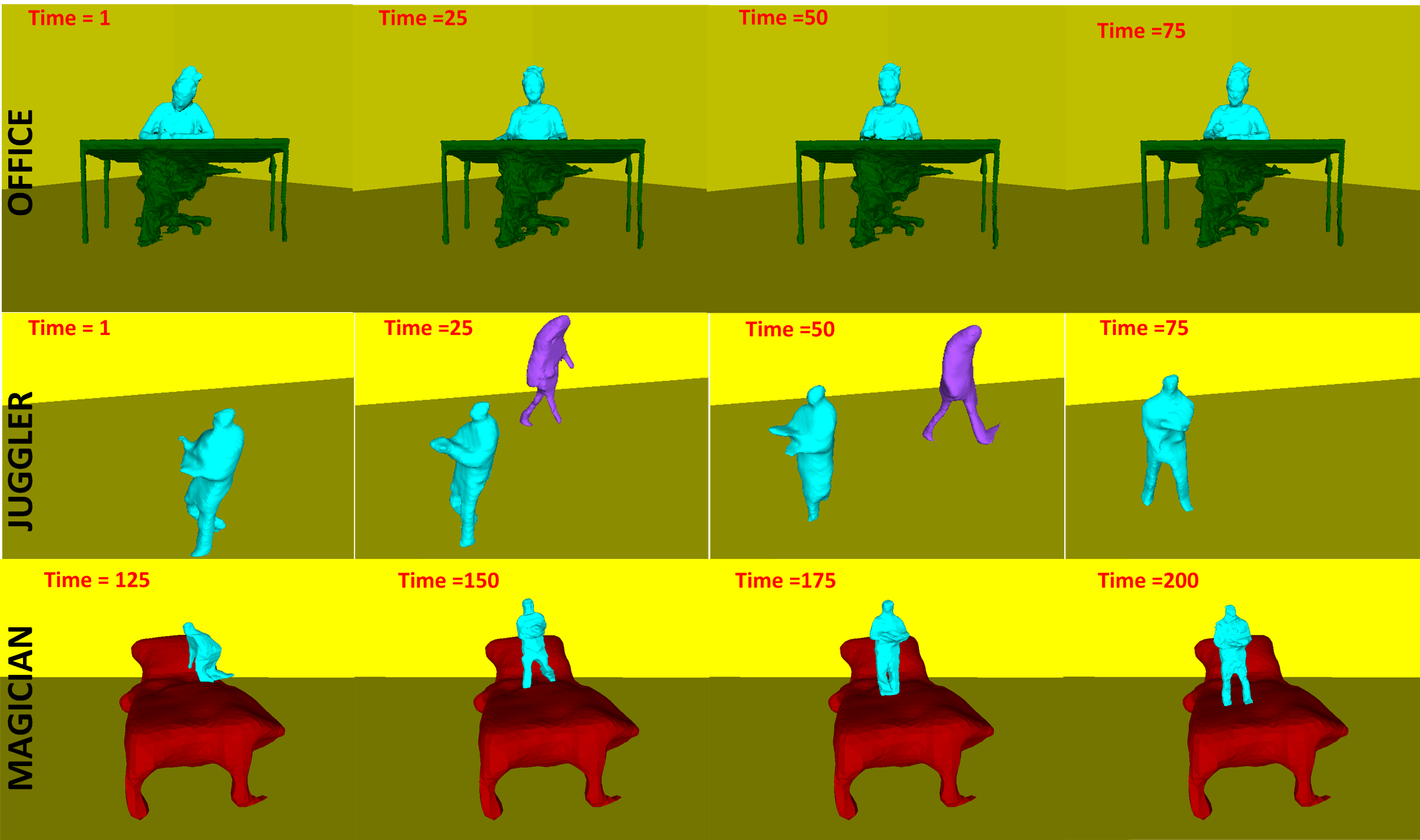}
		\caption{Complete scene reconstruction with 4D mesh sequence.}
		\label{fig:CompleteReconst}
	\end{center}
\end{figure}

%
\begin{figure}
	\begin{center}
		\includegraphics[width = 0.99\linewidth]{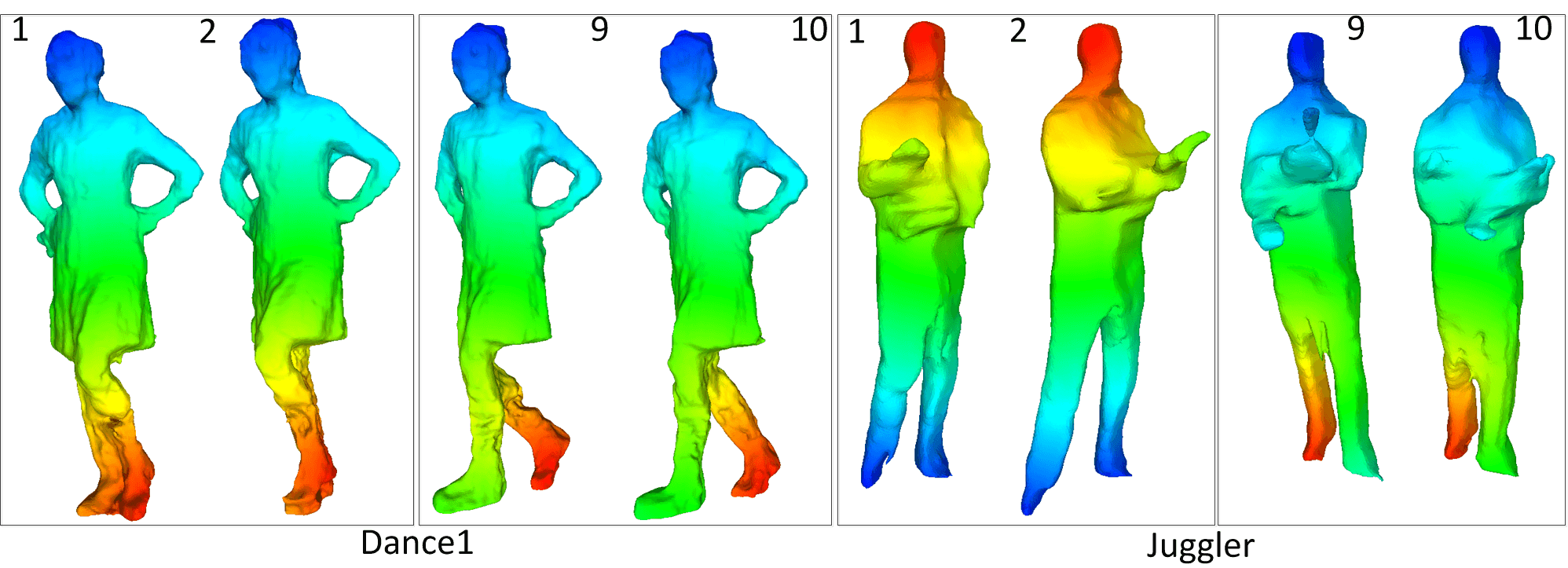}
		\caption{Frame-to-frame temporal alignment for Dance1 and Juggler dataset}
		\vspace{-0.3cm}
		\label{fig:temporal}
	\end{center}
\end{figure}
%
\subsection{Reconstruction evaluation}
Reconstruction results obtained using the proposed method are compared against Mustafa \cite{MustafaICCV15}, Guillemaut \cite{Guillemaut3dv}, and Furukawa \cite{Furukawa2010} for dynamic sequences. Furukawa \cite{Furukawa2010} is a per-frame multi-view wide-baseline stereo approach  which ranks highly on the middlebury benchmark \cite{Seitz06}  but does not refine the segmentation. 

The depth maps obtained using the proposed approach are compared against Mustafa and Guillemaut in Figure \ref{fig:depthMagnJug}. The depth map obtained using the proposed approach are smoother with low reconstruction noise compared to the state-of-the-art methods. Figure \ref{fig:meshes} and \ref{fig:dance2} present qualitative and quantitative comparison of our method with the state-of-the-art approaches.

Comparison of reconstructions demonstrates that the proposed method gives consistently more complete and accurate models. The colour maps highlight the quantitative differences in reconstruction.
As far as we are aware no ground-truth data exist for dynamic scene reconstruction from real multi-view video. 
In Figure \ref{fig:dance2} we present a comparison with the reference mesh available with the Dance2 dataset reconstructed using a visual-hull approach. This comparison demonstrates improved reconstruction of fine details with the proposed technique.

In contrast to all previous approaches the proposed method gives temporally coherent 4D model reconstructions with dense surface correspondence over time. The introduction of temporal coherence constrains the reconstruction in regions which are ambiguous on a particular frame such as the right leg of the juggler in Figure \ref{fig:meshes} resulting in more complete shape. Figure \ref{fig:CompleteReconst} shows three complete scene reconstructions with 4D models of multiple  objects. The Juggler and Magician sequences are reconstructed from moving handheld cameras.
\\
\\
\textbf{Computational Complexity:}
Computation times for the proposed approach vs other methods are presented in Table \ref{time}. 
The proposed approach to reconstruct temporally coherent 4D models is comparable in computation time to per-frame multiple view reconstruction and gives a $\sim$50\% reduction in computation cost compared to previous joint segmentation and reconstruction approaches using a known background. This efficiency is achieved through improved per-frame initialisation based on temporal propagation and the introduction of the geodesic star constraint in joint optimisation. Further results can be found in the supplementary material.\\
\\
\textbf{Temporal coherence:}
A frame-to-frame alignment is obtained using the proposed approach as shown in Figure \ref{fig:temporal} for Dance1 and Juggle dataset. The meshes of the dynamic object in Frame 1 and Frame 9 are color coded in both the datasets and the color is propagated to the next frame using the dense temporal coherence information. The color in different parts of the object is retained to the next frame as seen from the figure. The proposed approach obtains sequential temporal alignment which drifts with large movement in the object, hence successive frames are shown in the figure. 
\\
\\
{\bf Limitations:} As with previous dynamic scene reconstruction methods the proposed approach has a number of limitations: persistent ambiguities in appearance between objects will degrade the improvement achieved with temporal coherence; scenes with a large number of inter-occluding dynamic objects will degrade performance; the approach requires sufficient wide-baseline views to cover the scene. Background reconstruction is limited to coarse reconstruction of orthogonal planes based on the Manhattan world assumption.
%
\begin{figure}
	\begin{center}
		\includegraphics[width =0.99\linewidth]{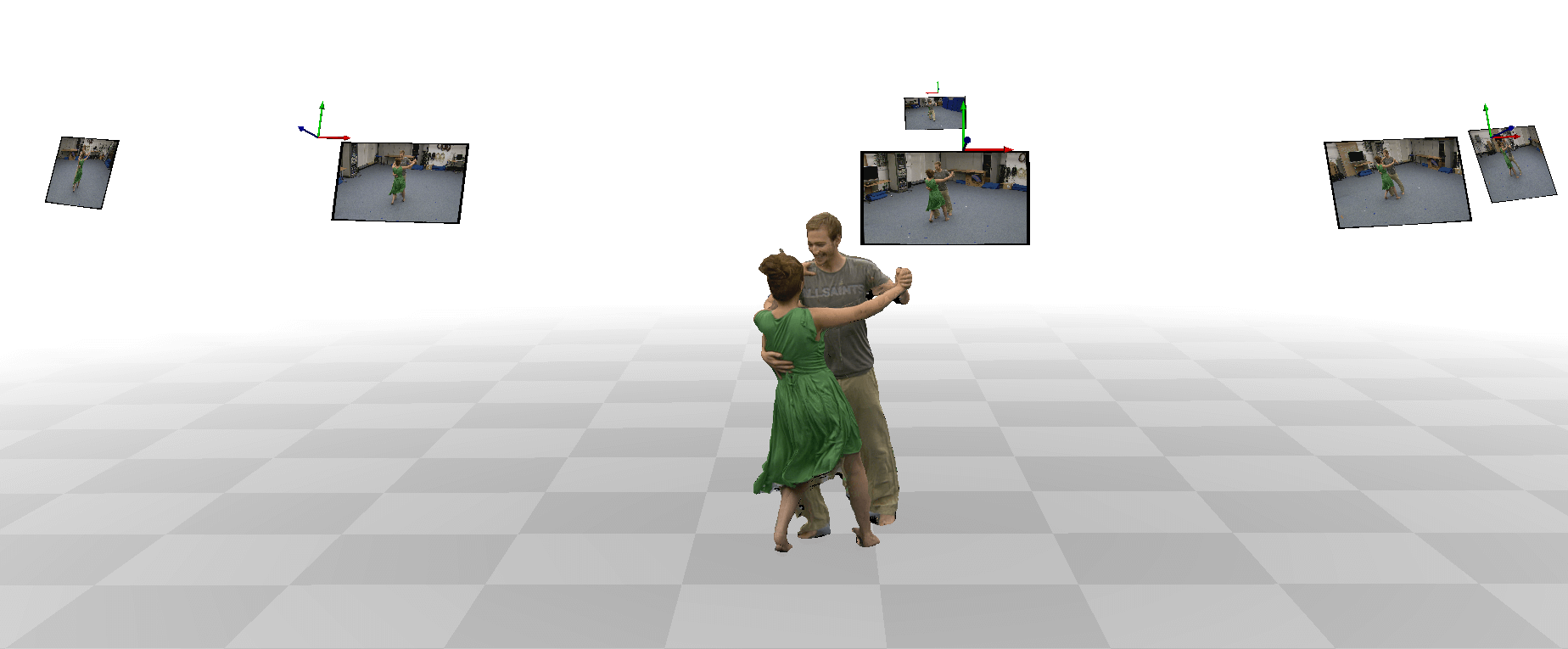}
		\caption{Application of proposed method for freeview-point video for Dance2 dataset.}
		\label{fig:fvvDance2}
	\end{center}
\end{figure}
%
\begin{figure*}
	\begin{center}
		\includegraphics[width = 0.90\linewidth]{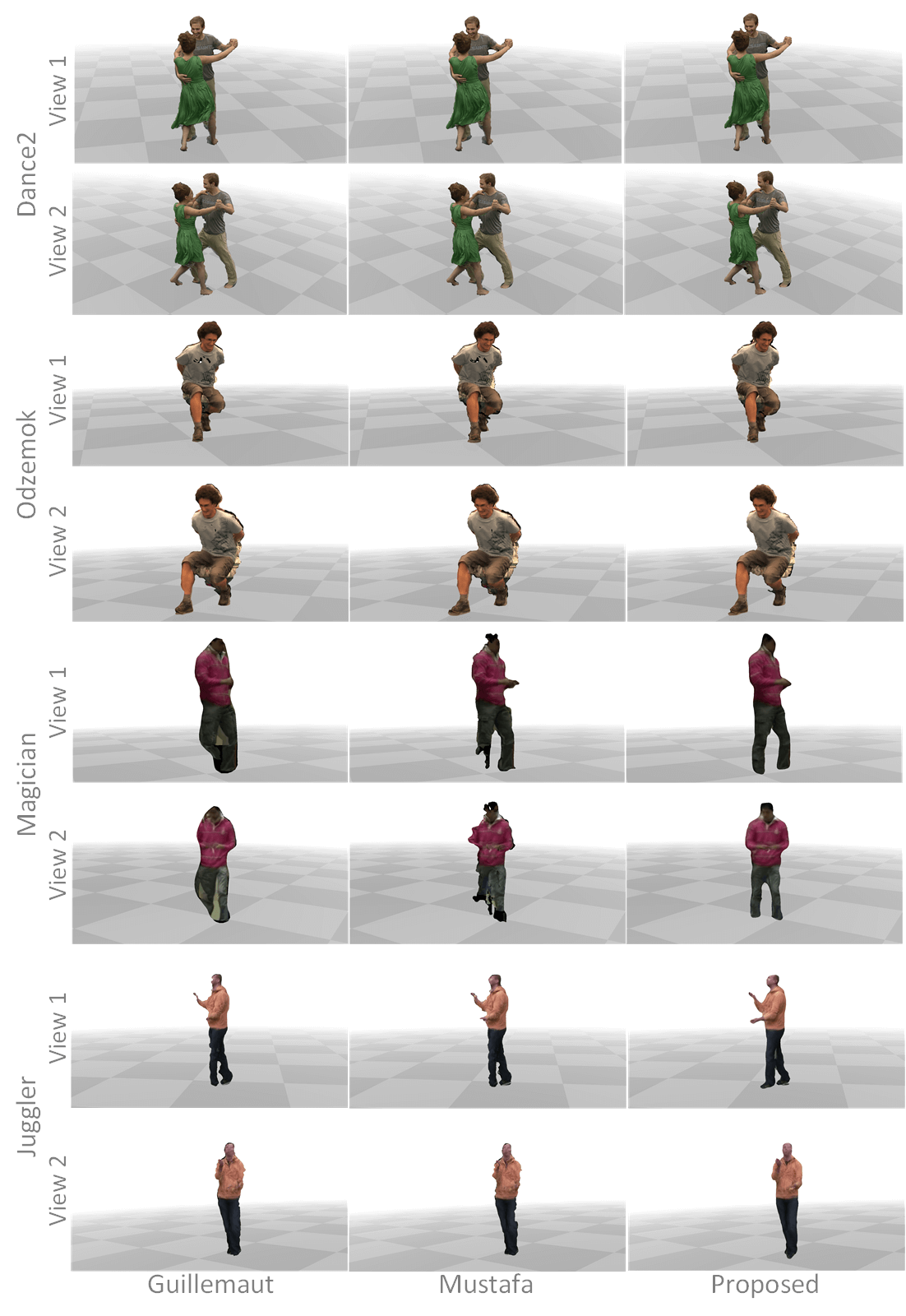}
		\caption{Comparison of Free-viewpoint rendering of proposed method against Mustafa and Guillemaut for 4 datasets.}
		\label{fig:fvvCompare}
	\end{center}
\end{figure*}
%
\begin{figure*}
	\begin{center}
		\includegraphics[width = 0.99\linewidth]{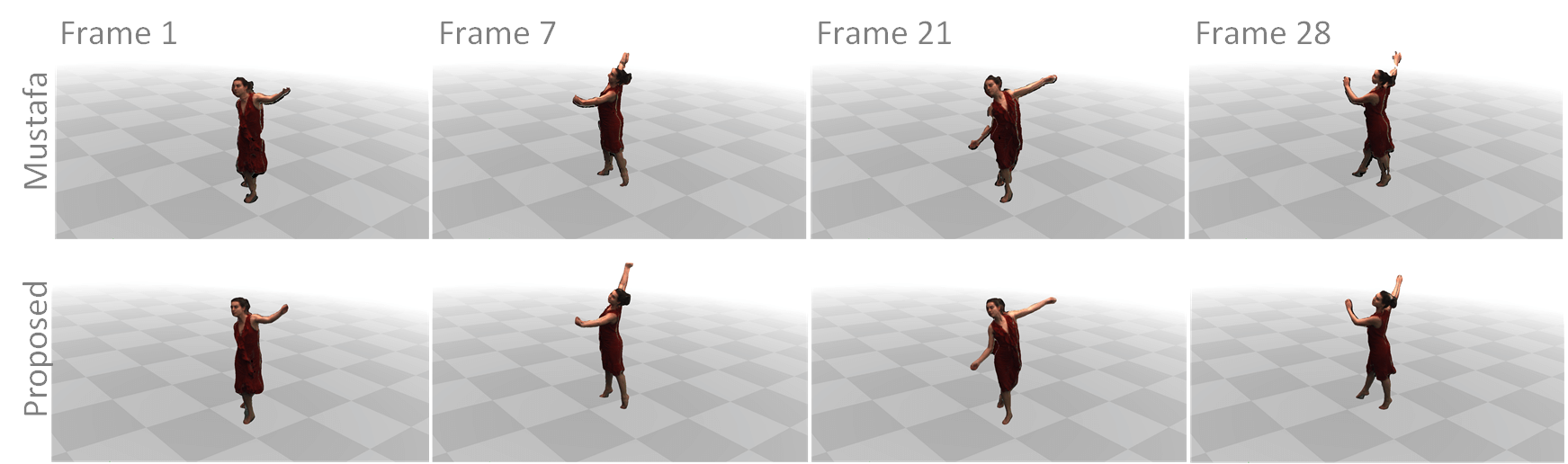}
		\caption{Comparison of Free-viewpoint rendering of proposed method against Mustafa for Dance1 sequence.}
		\vspace{-0.3cm}
		\label{fig:fvvCompareSeq}
	\end{center}
\end{figure*}
%
\begin{figure*}
	\begin{center}
		\includegraphics[width =0.99\linewidth]{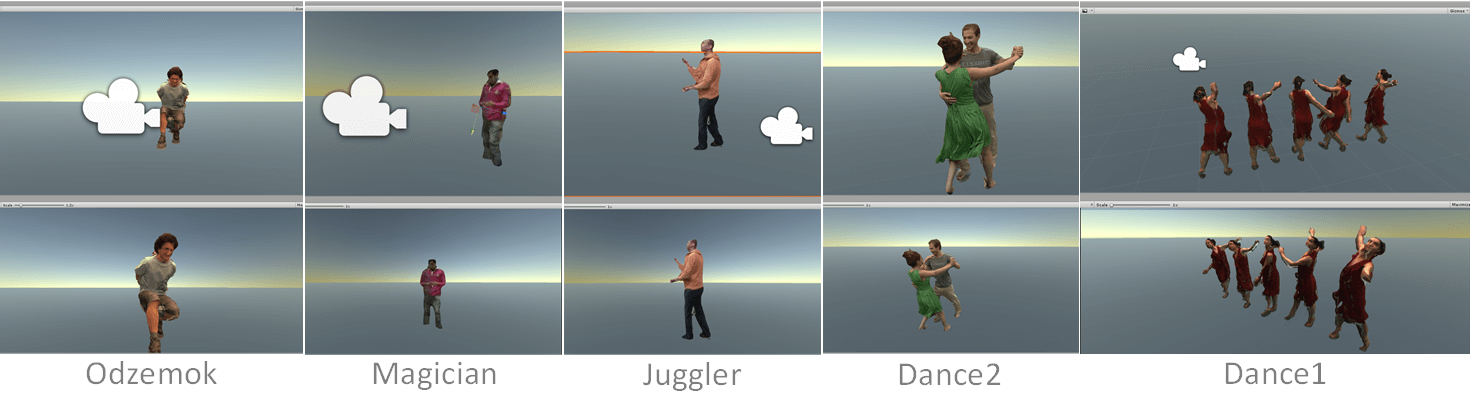}
		\caption{Application of proposed method for Virtual Reality. Renderings in Unity are shown for five datasets, including a sequence for Dance1.}
		\vspace{-0.5cm}
		\label{fig:VR}
	\end{center}
\end{figure*}
%
\begin{figure*}
	\begin{center}
		\includegraphics[width =0.99\linewidth]{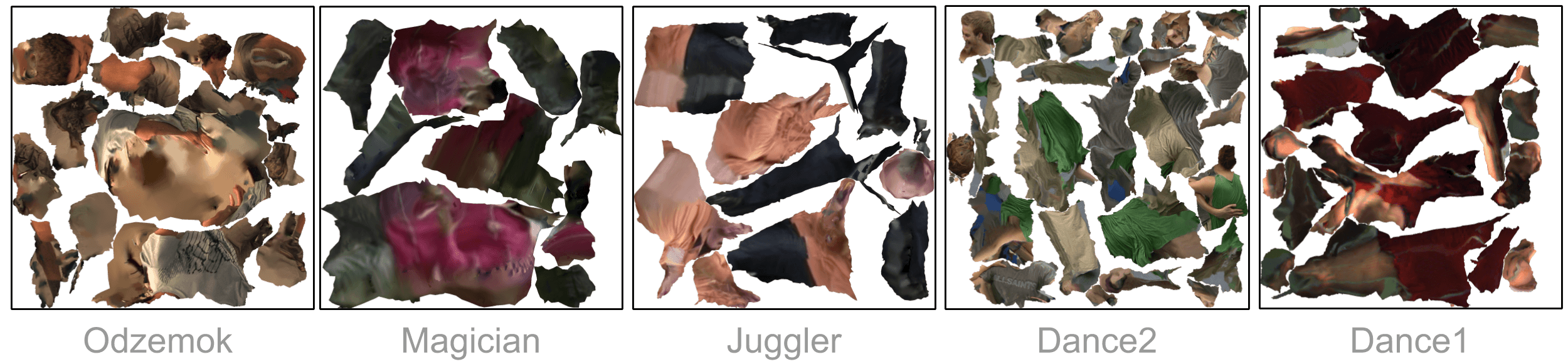}
		\caption{UV texture atlases for different dynamic datasets to render in VR at a single time instance.}
		\vspace{-0.7cm}
		\label{fig:VRtextures}
	\end{center}
\end{figure*}
%
\section{Applications}
\label{sec:applications}
The 4D meshes generated from the proposed approach can be used for applications in immersive content production such as FVV rendering and VR. Unlike the previous methods proposed framework does not require strong prior assumptions and manual interactions to create 4D meshes for real-world applications. This section demonstrates the results of these applications.
\subsection{Free-viewpoint rendering}
In FVV, the virtual viewpoint is controlled interactively by the user. The appearance of the reconstruction is sampled and interpolated directly from the captured camera images using cameras located close to the virtual viewpoint \cite{FVV}.

The proposed joint segmentation and reconstruction framework generates per-view silhouettes and a temporally coherent 4D reconstruction at each time instant of the input video sequence. This representation of the dynamic sequence is used for FVV rendering. To create FVV, a view-dependent surface texture is computed based on the user selected virtual view. This virtual view is obtained by combining the information from camera views in close proximity to the virtual viewpoint \cite{FVV}. FVV rendering gives user the freedom to interactively choose a novel viewpoint in space to observe the dynamic scene and reproduces fine scale temporal surface details, such as the movement of hair and clothing wrinkles, that may not be modelled geometrically. An example of a reconstructed scene and the camera configuration is shown in Figure \ref{fig:fvvDance2}.

A qualitative evaluation of images synthesised using FVV is shown in Figure \ref{fig:fvvCompare} and \ref{fig:fvvCompareSeq}. These demonstrate reconstruction results rendered from novel viewpoints from the proposed method against Mustafa \cite{Mustafa16} and Guillemaut \cite{Guillemaut2010} on publicly available datasets. This is particularly important for wide-baseline camera configurations where this technique can be used to synthesize intermediate viewpoints where it may not be practical or economical to physically locate real cameras. 
%
\subsection{Virtual reality rendering}
There is a growing demand for photo-realistic content in the creation of immersive VR experiences. The 4D temporally coherent reconstructions of the dynamic scenes obtained using the proposed approach enables the creation of photo-realistic digital assets that can be incorporated into VR environments using game engines such as Unity and Unreal Engine, as shown in Figure \ref{fig:VR} for single frame of four datasets and for a series of frames for Dance1 dataset.

In order to efficiently render the reconstructions in a game engine for applications in VR, a UV texture atlas is extracted using the 4D meshes from the proposed approach as a geometric proxy. The UV texture atlas at each frame are applied to the models at render time in unity for viewing in a VR headset. A UV texture atlas is constructed by projectively texturing and blending multiple view frames onto a 2D unwrapped UV texture atlas, see Figure \ref{fig:VRtextures}. This is performed once for each static object and at each time instance for dynamic objects allowing efficient storage and real-time playback of static and dynamic textured reconstructions within a VR headset.
%
\section{Conclusion}
This paper introduced a novel technique to automatically segment and reconstruct dynamic scenes captured from multiple moving cameras in general dynamic uncontrolled environments without any prior on background appearance or structure. The proposed automatic initialization was used to identify and initialize the segmentation and reconstruction of multiple objects. 
A framework for temporally coherent 4D model reconstruction of dynamic scenes from a set of wide-baseline moving cameras. The approach gives a complete model of all static and dynamic non-rigid objects in the scene. Temporal coherence for dynamic objects addresses limitations of previous per-frame reconstruction giving improved reconstruction and segmentation together with dense temporal surface correspondence for dynamic objects.
A sparse-to-dense approach is introduced to establish temporal correspondence for non-rigid objects using robust sparse feature matching to initialise dense optical flow providing an initial segmentation and reconstruction. Joint refinement of object reconstruction and segmentation is then performed using a multiple view optimisation with a novel geodesic star convexity constraint that gives improved shape estimation and is computationally efficient.
Comparison against state-of-the-art techniques for multiple view segmentation and reconstruction demonstrates significant improvement in performance for complex scenes. The approach enables reconstruction of 4D models for complex scenes which has not been demonstrated previously.
%
\begin{acknowledgements}
	This research was supported by the Royal Academy of Engineering Research Fellowship RF-201718-17177 and the EPSRC Platform Grant on Audio-Visual Media Research EP/P022529.
\end{acknowledgements}
\bibliographystyle{spmpsci}      
\bibliography{Bibliography}

\end{document}